\documentclass[letterpaper]{article} 
\usepackage{aaai25}  
\usepackage{times}  
\usepackage{helvet}  
\usepackage{courier}  
\usepackage[hyphens]{url}  
\usepackage{graphicx} 
\usepackage{natbib}  
\usepackage{caption} 
\usepackage{algorithm}
\usepackage{algorithmic}

\usepackage{newfloat}
\usepackage{listings}
\DeclareCaptionStyle{ruled}{labelfont=normalfont,labelsep=colon,strut=off} 
\floatstyle{ruled}
\newfloat{listing}{tb}{lst}{}
\floatname{listing}{Listing}
\usepackage{soul}
\usepackage{cancel}
\usepackage{xcolor}
\usepackage{booktabs}
\usepackage{amsmath}
\usepackage{graphicx}
\usepackage{kotex}
\usepackage{colortbl}
\usepackage{amsfonts} 

\usepackage{tikz}
\usepackage{algorithm}
\usepackage{algorithmic}
\usepackage{newfloat}
\usepackage{listings}
\usepackage{soul}
\usepackage{cancel}
\usepackage{xcolor}
\usepackage{booktabs}
\usepackage{amsmath}
\usepackage{graphicx}
\usepackage{kotex}
\usepackage{colortbl}
\usepackage{amsfonts} 

\usepackage{tikz}
\usepackage{algorithm}
\usepackage{algorithmic}

\usepackage{newfloat}
\usepackage{listings}
\usepackage{soul}
\usepackage{cancel}
\usepackage{xcolor}
\usepackage{booktabs}
\usepackage{amsmath}
\usepackage{graphicx}
\usepackage{kotex}
\usepackage{amsfonts} 
\usepackage{colortbl}
\usepackage{tikz}
\usepackage{booktabs}  
\usepackage{tabularx}  
\usepackage{multirow} 

\title{DEEPTalk: Dynamic Emotion Embedding for Probabilistic Speech-Driven 3D Face Animation}
\author{Jisoo Kim\textsuperscript{\rm 1}\equalcontrib,
    Jungbin Cho\textsuperscript{\rm 1}\equalcontrib,
    Joonho Park\textsuperscript{\rm 2},
    Soonmin Hwang\textsuperscript{\rm 1},
    Da Eun Kim\textsuperscript{\rm 2},
    Geon Kim\textsuperscript{\rm 2},\\
    Youngjae Yu\textsuperscript{\rm 1}\thanks{Corresponding author}}
\affiliations {
    \textsuperscript{\rm 1}Yonsei University\\
    \textsuperscript{\rm 2}GIANTSTEP Inc.\\
    \{jisoo6687, whwjdqls99, smsm0307, yjy\}@yonsei.ac.kr\\
    \{joonho.park, daeun.kim, geon.kim\}@giantstepcorp.com
}

\usepackage{bibentry}

\begin{document}
\frenchspacing

\maketitle

\begin{abstract}
Speech-driven 3D facial animation has garnered lots of attention thanks to its broad range of applications. Despite recent advancements in achieving realistic lip motion, current methods fail to capture the nuanced emotional undertones conveyed through speech and produce monotonous facial motion. These limitations result in blunt and repetitive facial animations, reducing user engagement and hindering their applicability. To address these challenges, we introduce DEEPTalk, a novel approach that generates diverse and emotionally rich 3D facial expressions directly from speech inputs. To achieve this, we first train DEE (Dynamic Emotion Embedding), which employs probabilistic contrastive learning to forge a joint emotion embedding space for both speech and facial motion. This probabilistic framework captures the uncertainty in interpreting emotions from speech and facial motion, enabling the derivation of emotion vectors from its multifaceted space. Moreover, to generate dynamic facial motion, we design TH-VQVAE (Temporally Hierarchical VQ-VAE) as an expressive and robust motion prior overcoming limitations of VAEs and VQ-VAEs. Utilizing these strong priors, we develop DEEPTalk, a talking head generator that non-autoregressively predicts codebook indices to create dynamic facial motion, incorporating a novel emotion consistency loss. Extensive experiments on various datasets demonstrate the effectiveness of our approach in creating diverse, emotionally expressive talking faces that maintain accurate lip-sync. Our project page is available at {\scriptsize\texttt{https://whwjdqls.github.io/deeptalk\_website/}}
\end{abstract}

\section{Introduction}

Speech-driven 3D facial motion generation has a wide range of applications, encompassing avatar animation for game or cinematic productions, virtual chatbots, and immersive virtual meetings within virtual reality environments. Despite substantial advancements in accurate lip synchronization with speech, as demonstrated by recent research \cite{richard2021meshtalk,fan2022faceformer, xing2023codetalker, stan2023facediffuser}, most of these methods still produce blunt and unexpressive facial expressions. Since facial expressions are crucial for conveying nonverbal cues, this limitation significantly reduces their effectiveness in scenarios requiring realistic interactions, such as interactions with non-player characters in games or emotionally responsive virtual chatbots. Therefore, it is essential to focus on enhancing the full range of facial expressions, not just the lip movements. 

\begin{figure}[t]
\includegraphics[width=0.99\linewidth]{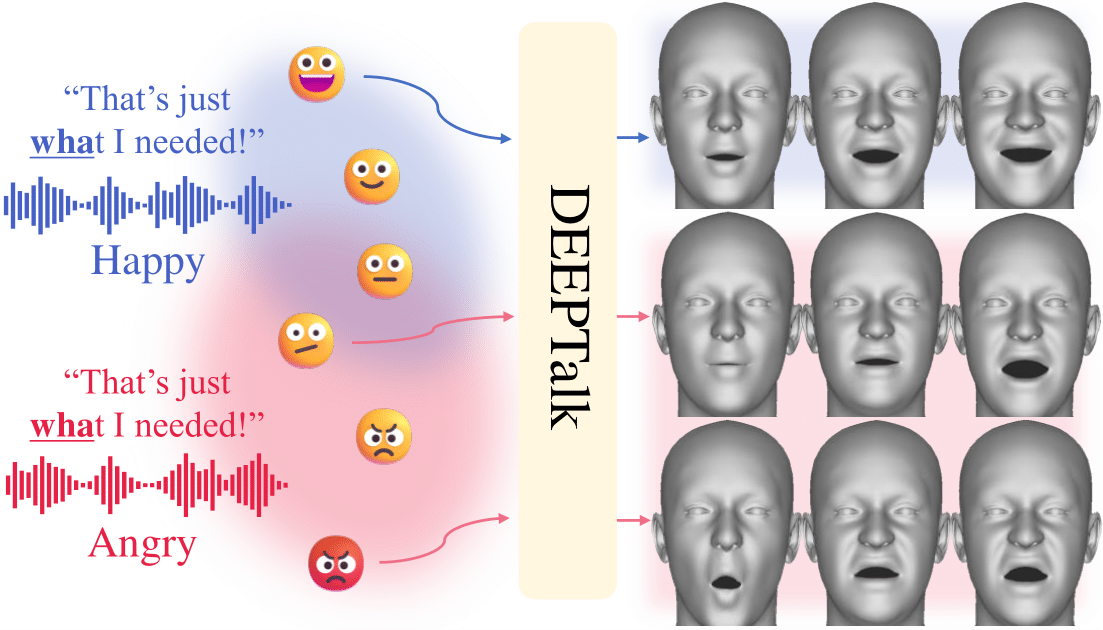}
\caption{Overview of DEEPTalk. Starting with an emotional speech input (left), we extract probabilistic emotion embeddings (depicted as blobs), and sample from these embeddings to generate diverse emotional facial animations aligned with the input speech (right).}
\label{fig:main_figure}
\end{figure}

Previous studies on talking faces have generated expressions using either emotion labels \cite{danvevcek2023emotional, gan2023emo2d, pan2023emo2d, ji202emo2d} or reference expressions \cite{ji2022eammemo2d_reference, tan2024say, ma2023styletalk_emo2d_reference, liang2022expressive_emo2d_reference}. Using emotion labels provides expressive but limited outcomes, while reference expressions offer more diversity at the cost of needing numerous expressive references. Moreover, both approaches fail to capture vocal nuances, often leading to misalignment between facial expressions and speech. As shown in Figure~\ref{fig:main_figure}, the phrase "That's just what I needed!" can carry different meanings based on the emotion it conveys (e.g., happiness or anger), emphasizing the importance of facial expressions that reflect the intended sentiment. Without this alignment, expressions may appear unnatural from spoken words, contributing to uncanny valley effect \cite{uncannyvalley}.

Therefore, the most effective approach to generating nonverbal facial expressions from speech is to leverage the rich information embedded within the speech, which simultaneously conveys the speaker's intentions and emotions. Central to this process is prosody, encompassing non-linguistic elements such as pitch, speed, volume, and tone variations. Prosody is crucial due to its intricate link with facial expressions as shown in \cite{prosody_and_face}\cite{prosody_and_face2}.

Utilizing this groundwork, we propose a talking head model, DEEPTalk, that leverages both emotion and motion priors to generate diverse emotional 3D facial expressions directly from speech inputs. 
We first utilize cross-modal contrastive learning to capture the emotional correlation between speech and 
facial expressions. Unlike \cite{crossmodal_transfer_inthewild}, which used this correlation primarily for emotion recognition from unlabeled data, we aim to use it to develop a joint embedding space, which we call DEE(Dynamic Emotion Embedding). Given the inherent ambiguity in both speech and facial expressions—where multiple expressions can correspond to a single piece of speech—we utilize probabilistic embeddings \cite{pcem, chun2023improved}. These embeddings are designed to model the uncertainty associated with the inputs and facilitate sampling from the probability distribution. This allows the generation of diverse emotional faces from the same speech input as illustrated by the red lines in Figure~\ref{fig:main_figure}.
After constructing a joint embedding space for facial motion and speech, we use it as a strong emotion prior to train an emotional talking head model. 

We then aim to build an expressive motion prior that is robust to perceptual losses. Recent studies have demonstrated that by learning motion priors from discrete codebooks, it is possible to generate a diverse range of facial and body motions \cite{li2021audio2gestures, talkshow, xing2023codetalker, learning2listen_motionprior}. However, due to the dynamic nature of emotional talking faces, VQ-VAE\cite{van2017neural} alone struggles to capture the entire motion space fully. Recognizing that facial motion encompasses different temporal hierarchies—for instance, the mouth region exhibits high frequencies while the upper face displays lower temporal frequencies—we train a hierarchical discrete motion prior to effectively address these variations effectively.

Building on the aforementioned robust emotion and motion priors, DEEPTalk is specifically engineered to non-autoregressively map emotional speech to our codebook indices. To ensure that generated expressions consistently reflect the input speech emotions, we introduce a novel emotion consistency loss. Training incoporates Gumbel-Softmax \cite{jang2016categorical} and differentiable rendering to ensure end-to-end differentiability. Our qualitative and quantitative assessments, including extensive user studies, demonstrate that DEEPTalk excels at generating diverse emotional facial motions while also outperforming lip synchronization.

In summary, our main contributions are as follows: we design a Dynamic Emotion Embedding (DEE) that jointly learns from speech and facial motion sequences through probabilistic contrastive learning. Additionally, we propose a novel temporally hierarchical VQ-VAE (TH-VQVAE) to construct a motion prior that is both expressive and robust. Finally, we train a talking head model, DEEPTalk, leveraging these strong priors to generate diverse and expressive emotional facial motions while also outperforming state-of-the-art models on lip synchronization.

\section{Related Work}

\subsubsection{Speech-Driven 3D Face Animation.}
The availability of large 4D mesh datasets synchronized with audio \cite{richard2021meshtalk,fanelli20103} has greatly advanced deep learning-based 3D face animation, enabling robust lip synchronization. Extending these advancements, VOCA \cite{cudeiro2019capture} improves realism by generating animations from any speech inputs, employing time convolutions with a speaker identity one-hot vector. FaceFormer \cite{fan2022faceformer} uses a transformer-based model to generate facial movements auto-regressively. However, despite accurate lip synchronization, the generated upper face remains static as it is less correlated with input speech. MeshTalk \cite{richard2021meshtalk} addresses this by disentangling audio-correlated and uncorrelated movements using a categorical latent space to model upper face dynamics. Similarly, CodeTalker \cite{xing2023codetalker} utilizes a discrete motion prior with VQ-VAE to reconstruct dynamic facial motions across the entire face. However, these methods generate deterministic motion, overlooking the inherently non-deterministic relationship between facial motion and speech. Therefore, FaceDiffuser \cite{stan2023facediffuser} addresses the limitation of deterministic models by employing a diffusion model, which is probabilistic in nature, to generate multiple facial motions for a given speech. However, it still falls short in capturing the diverse emotional expressions conveyed through speech, which is crucial for enhancing interactiveness for talking face models.

  \begin{figure*}[ht]
    \centering
    \includegraphics[width=0.99\linewidth]{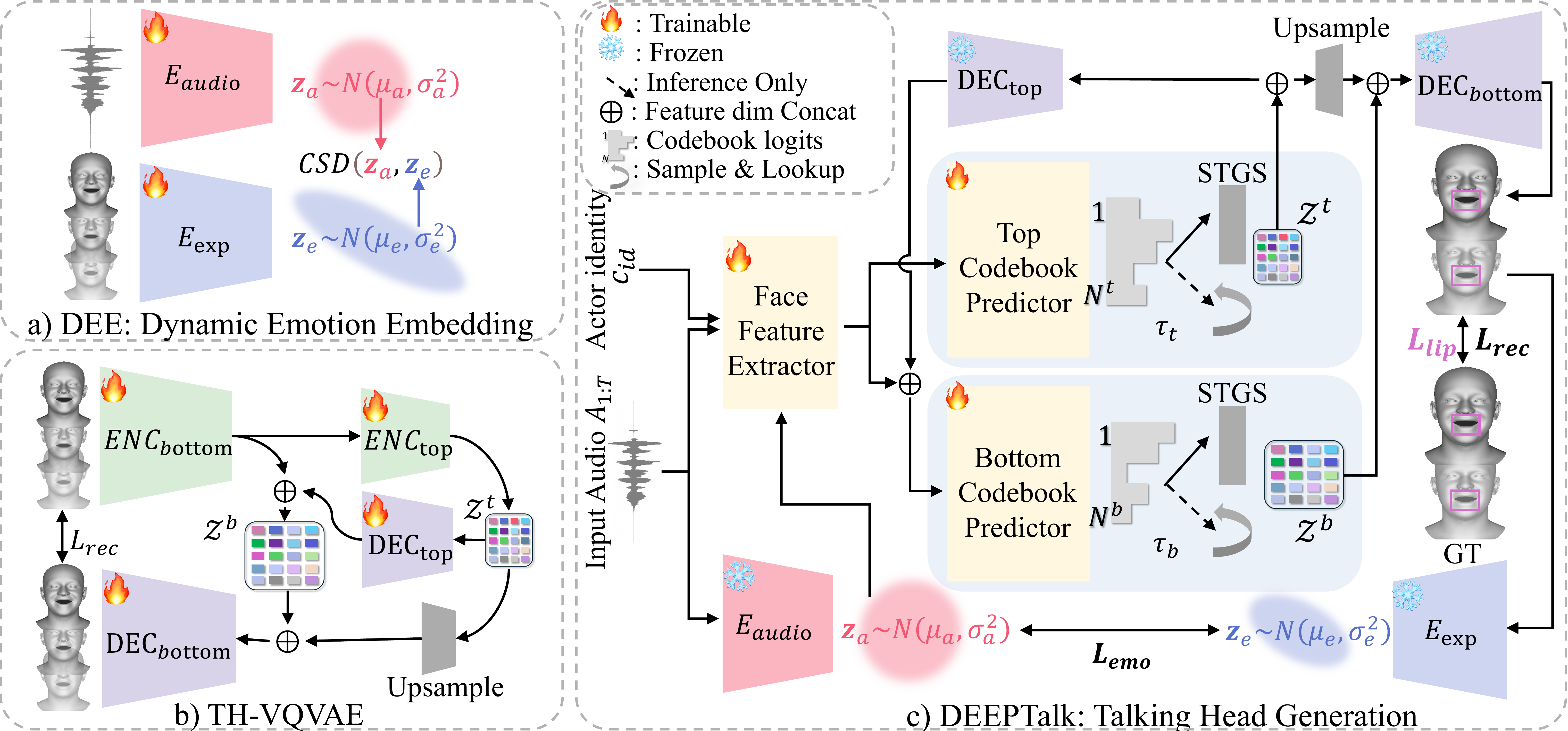}
    \caption{Overall Architecture of Our Method. (a) $E_{audio}$ and $E_{exp}$ are trained to predict mean and variance for a joint audio-facial emotion embedding space, DEE. (b) We train TH-VQVAE with separate codebooks, $\mathcal{Z}^b$ and $\mathcal{Z}^t$, for low and high-frequency motions, respectively. (c) DEEPTalk first extracts face features, predict top and bottom codebook indices, and use frozen TH-VQVAE decoders to decode the quantized motion features. To ensure emotion alignment between input audio and the predicted facial expressions, we introduce an emotional consistency loss $L_{emo}$ by utilizing DEE.}
    \label{fig:DEETPTalkfig}
\end{figure*}

\subsubsection{Emotional 3D Face Animation.}
Recent studies have underscored the critical role of emotion in creating realistic and expressive 3D facial motions by integrating additional emotional information. Specifically, EMOTE \cite{danvevcek2023emotional} employs one-hot labels to control emotion, producing emotional facial motions through an emotion-content disentanglement loss. Chen et al. \cite{chen2023expressive} utilized the logits from an emotion classifier applied to reference images as an emotion prior during training, generating emotional facial motion through an emotion-augmented network. However, these methods require explicit control, lacking a direct connection to the emotion conveyed in the actual speech.
 The work most related to ours is EmoTalk \cite{peng2023emotalk}, which aims to generate emotional blendshapes from audio input only, utilizing an emotion-content disentangling method. However, the training of EmoTalk's emotion-content disentangling encoder is constrained to specific datasets like RAVDESS \cite{RAVDESS}, containing the same sentences expressed in multiple emotions. This constraint limits the use of larger, more diverse datasets such as MEAD \cite{kaisiyuan2020mead}, ultimately hindering the model's ability to generalize emotional expressions on other datasets.
Unlike previous methods that either disentangle content and emotion from speech or rely on explicit conditioning, our approach establishes a robust emotion prior by leveraging the natural correlation between speech and facial expressions, enabling seamless integration into the talking face training pipeline.

\section{Method}
\subsubsection{Overview}
Our goal is to synthesize emotional facial motion solely from speech. However, the complex many-to-many relationship between speech and facial expressions poses challenges for generating expressive emotional faces. To overcome this, we build (1) DEE, a joint probabilistic emotional space to encode both audio and facial motion. DEE sample emotion embeddings from audio, and enable generated facial motions aligned to the corresponding embedding. Also for robust and expressive motion prior, we design (2) TH-VQVAE to learn a discrete motion space capable of capturing both high and low-frequency motions.
Building on these priors, we train (3) DEEPTalk to map emotional audio to motion codebooks, resulting in facial animations that are both emotionally expressive and accurately lip-synced. Additional details on each component are provided in the Supplementary.
\subsubsection{Formulation.} Our task can be formulated as follows: Let $\boldsymbol{A}_{1:T} = ( \boldsymbol{a}_1, \dots, \boldsymbol{a}_T)$ be a sequence of input speech snippets where each $\boldsymbol{a}_t \in \mathbb{R}^{D}$ has $D$ sampled audio, and let $\boldsymbol{F}_{1:T} = (\boldsymbol{f}_1, \cdots, \boldsymbol{f}_T)$, where $\boldsymbol{f}_t \in \mathbb{R}^{d_{exp} + 3}$ is a sequence of FLAME expression parameters $\boldsymbol{\Psi}_t \in \mathbb{R}^{d_{exp}}$ concatenated with jaw parameters $\boldsymbol{\theta}^\text{jaw}_t \in \mathbb{R}^{3}$.
\begin{equation}
  \boldsymbol{f}_t = [\boldsymbol{\Psi}_t,\boldsymbol{\theta}^\text{jaw}_t ],
\end{equation}
Our goal is to analyze the content and emotion of $\boldsymbol{A}_{1:T}$ and, together with the one-hot speaker identity $\boldsymbol{c}_{id}$, predict FLAME expression parameters $\boldsymbol{\hat{F}}_{1:T} = (\boldsymbol{\hat{f}}_1, \cdots, \boldsymbol{\hat{f}}_T)$ that align with the input speech. As our talking head model is non-auto-regressive, denoting $\theta$ as model parameters, our end-to-end procedure can be written as
\begin{equation}
\hat{\boldsymbol{F}}_{1:T}=\operatorname{DEEPTalk}_\theta(\boldsymbol{A}_{1:T}, \boldsymbol{c}_{id}),
\end{equation}

\subsection{DEE: Dynamic Emotional Embedding}
\label{sec:DEE}
\subsubsection{Emotion feature extraction.} We utilized the recently proposed emotion2vec~\cite{ma2023emotion2vec} to extract emotion features in our audio encoder. Denoting feature extractor as $F_{audio}$, this process is defined as :
\begin{equation}
  F_{audio}(\boldsymbol{A}_{1:T}) \rightarrow \epsilon_{audio}
\end{equation}
 For the expression feature extractor in the expression encoder, we first trained an emotion classification model using the AffectNet dataset~\cite{mollahosseini2017affectnet}. As Affectnet is an image dataset, we applied a 3D flame parameter reconstruction method~\cite{danvevcek2022emoca} to generate 3D pseudo ground truth for each image in AffectNet and trained an emotion recognition model.  The trained encoder was then repurposed as the facial expression feature extractor $F_{exp}$. This process is defined as:
\begin{equation}
  F_{exp}(\boldsymbol{F}_{1:T}) \rightarrow \epsilon_{exp}
\end{equation}
\subsubsection{Emotional space construction.} 
With both audio and expression feature extractors in place, DEE trains the audio and expression encoders separately. The overview of DEE is illustrated in Figure~\ref{fig:DEETPTalkfig}(a). Each encoder $E$ comprises two distinct heads, followed by a Generalized Pooling Operator (GPO) \cite{GPO}, with one head dedicated to $\mu$ and the other to $\log\sigma^2$, similar to the approach in PCME++\cite{chun2023improved}. The final probabilistic emotion embeddings for audio and expression can be written as follows:
\begin{equation}
  E_{audio}(\epsilon_{audio}) \rightarrow Z_{a} \sim N(\mu_{a}, \sigma_{a}^2)
\end{equation}
\begin{equation}
  E_{exp}(\epsilon_{exp}) \rightarrow Z_{e} \sim N(\mu_{e}, \sigma_{e}^2)
\end{equation}

\subsubsection{Objectives.} We train DEE on the Closed-Form Sampled distance (CSD), between audio and expression probabilistic embeddings $Z_{a}, Z_{e}$:
\begin{equation}
  CSD(Z_{a}, Z_{e}) = \Vert \mu_{a}-\mu_{e} \Vert^{2}_{2} + \Vert \sigma_{a}^2+\sigma_{e}^2 \Vert_{1}
\end{equation}

\subsection{TH-VQVAE: Temporally Hierarchical VQ-VAE}
\label{sec:TH-VQVAE}
We addressed the challenge of modeling high-frequency lip movements and slower facial motions by extending VQ-VAE2\cite{VQVAE2} into the temporal motion domain, creating TH-VQVAE. This model uses distinct codebooks for different motion frequencies, enhancing reconstruction quality and the capabilities of the Talking Head Generator. It enables fine-grained lip movements and dynamic facial expressions while also offering controllability at each hierarchical level.

\subsubsection{Model Architecture. } TH-VQVAE consists of a bottom encoder $\text{ENC}_{bottom}$, a top encoder $\text{ENC}_{top}$, a bottom decoder $\text{DEC}_{bottom}$, a top decoder $\text{DEC}_{top}$, and two facial motion codebooks $\mathcal{Z}^{b}$ and $\mathcal{Z}^{t}$. Each codebook can be formulated as 
\begin{equation}
\mathcal{Z}^{b}=\left\{\mathbf{z}^{b}_k \in \mathbb{R}^{C^{b}}\right\}_{k=1}^{N^{b}},
\mathcal{Z}^{t}=\left\{\mathbf{z}^{t}_k \in \mathbb{R}^{C^{t}}\right\}_{k=1}^{N^{t}}
\end{equation}
where each represents fine and coarse facial motion. Any sequence of facial motion $\boldsymbol{F}_{1:T}$ can be represented by one item on each codebook $\boldsymbol{z}^b_i, \boldsymbol{z}^t_j$ and decoded through $\text{DEC}_{bottom}$ into the corresponding facial motion. As depicted in Figure~\ref{fig:DEETPTalkfig}(b), the input motion segment $x=\mathbf{F}_{1: T} \in \mathbb{R}^{T \times\left(d_{exp}+3\right)}$ is first encoded to a bottom motion feature and then encoded to top motion feature.
\begin{equation}
\hat{z}^b=\text{ENC}_{bottom}(x) \in \mathbb{R}^{\tau^{b} \times (C^{b}-C^{t})},
\end{equation}
where $\tau^{b}=\frac{T}{{q^{b}}}$ and $\tau^{t} =\frac{\tau}{{q^{t}}}$ is the length of the sequence divided by bottom quant factor $q^{b}$ and length of the sequence of the bottom motion feature divided by top quant factor $q^{t}$. Using the top motion feature, we obtain a top quantized sequence $z^{t}_q$, then use this to get decoded top features ${z}^{t}_d$ as
\begin{equation}
{z}^{t}_q=\arg \min _{{z}^t_t \in \mathcal{Z}^t} \left\|\hat{z}^t-z_t^t\right\| \in \mathbb{R}^{\tau^{t} \times C^{t}}.
\end{equation}
\begin{equation}
{z}^{t}_d=\text{DEC}_{top}({z}^{t}_q) \in \mathbb{R}^{\tau^{b} \times C^{t}},
\end{equation}
and stack $\hat{z}^b$ and ${z}^{t}_d$ along the feature dimension and obtain  ${z}^{b}_q$ by
\begin{equation}
{z}^{b}_q=\arg \min _{{z}^b_t \in \mathcal{Z}^b} \left\| [\hat{z}^b,{z}^{t}_d]-z_t^b\right\| \in \mathbb{R}^{\tau^{t} \times C^{b}}.
\end{equation}
Finally, we upsample ${z}^{t}_q$ from $\tau^t$ to match the temporal dimension $\tau^b$ and decode the stacked upsampled top quantized feature and bottom quantized feature to reconstruct the original facial motion inputs.   
\begin{equation}
\hat{x}=\text{DEC}_{bottom}\left([\text{Upsample}({z}^{t}_q),{z}^{b}_q ]\right)
\end{equation}
where $\hat{x}$ is a reconstruction of the input motion sequence $x$. We use the standard VQ-VAE loss to train TH-VQVAE. Details are included in the Supplementary.

\subsection{DEEPTalk: Talking Head Generator}

\begingroup
\setlength{\tabcolsep}{0.5mm}
\begin{table*}[t]
\centering
\fontsize{9}{9}\selectfont
\begin{tabular}{lccccccccccccccc}
    \toprule
    \textbf{Method} & \multicolumn{5}{c}{\textbf{CREMA-D}} & \multicolumn{3}{c}{\textbf{RAVDESS}} & \multicolumn{2}{c}{\textbf{HDTF}} & \multicolumn{5}{c}{\textbf{MEAD}} \\
    \cmidrule(lr){2-6} \cmidrule(lr){7-9} \cmidrule(lr){10-11} \cmidrule(lr){12-16}
    & FID$\downarrow$&  FFD$\downarrow$& Emo-FID$\downarrow$ & LSE-D$\downarrow$ & LSE-C$\uparrow$ & FID$\downarrow$ & FFD$\downarrow$& Emo-FID$\downarrow$ & LSE-D$\downarrow$ & LSE-C$\uparrow$ & FID$\downarrow$ & FFD$\downarrow$ & Emo-FID$\downarrow$ & LSE-D$\downarrow$ & LSE-C$\uparrow$ \\
    \midrule
    FaceFormer & 17.05 & \underline{62.88}& 27.69 & 8.659 & 1.460 & 27.43 & \underline{50.40} & 40.45 & \underline{11.88} & \underline{0.900} & 31.40  & \underline{78.62} & 35.67 & \underline{10.70} & \textbf{1.106} \\
    EmoTalk* & 64.75 & -& 149.1 & 9.132 & 1.413 & - & - & - &-& - & 88.33 & - & 228.9 & 10.77 & 0.896 \\
    FaceDiffuser & \underline{14.78} & 79.45 & \underline{23.24} & 8.686 & 1.507 & \underline{17.06}  & 81.74 & 31.23 & 12.48 & 0.515 & 30.93  & 119.2 & 68.30 & 10.93 & 0.784 \\
    EMOTE & 22.24 & 85.13 & 33.72 & \underline{8.612} & \textbf{1.573} & 20.22  & 62.05& \underline{27.40} & 12.38 & 0.491 & \underline{27.22} & 93.17 & \underline{16.83} & 10.73 & 1.035 \\
    Ours & \textbf{11.58} & \textbf{50.00}& \textbf{11.99} & \textbf{8.535} & \underline{1.523} & \textbf{11.94}  & \textbf{32.82}& \textbf{11.44} & \textbf{11.83} & \textbf{0.917} & \textbf{26.07} & \textbf{64.02} & \textbf{15.16} & \textbf{10.65} & \underline{1.103} \\
    \bottomrule
\end{tabular}
\caption{Quantitative Evaluation Results. Best performance in bold, and the second best \underline{underlined}. *EmoTalk does not predict FLAME parameters preventing evaluation of FFD and was trained on RAVDESS and HDTF. LSE-C reported in Supplementary.}
\label{tab:combined_results}
\end{table*}
\endgroup
\label{sec:DEEPTalk}

Shown in Figure~\ref{fig:DEETPTalkfig}(c), the Talking head generator is composed of two distinct modules. 1) Face Feature Extractor that extracts facial features from audio and emotion inputs, and 2) Codebook Predictor that predicts codebook indices given facial features. 

\subsubsection{Face Feature Extractor.}
In order to generate emotional face features aligned with input speech,  we employ Wav2Vec 2.0 \cite{baevski2020wav2vec} to encode content and DEE's audio encoder $E_{audio}$ to encode emotion. During inference, we can sample from the output distribution of $E_{audio}$ and control its uncertainty through scaling the $\log{\sigma^2}$ of $E_{audio}$ output with the uncertainty control factor $\alpha$. Both features are concatenated and fed into a transformer model to generate emotional face features. 

\subsubsection{Codebook Predictor.}
Instead of predicting each codebook's index at once, we utilize two separate models: the top codebook predictor and the bottom codebook predictor. First, we obtain low-frequency motion features and use them as a condition to predict high-frequency motion features. Specifically, the top codebook predictor first predicts the logit distribution of the top codebook and indexes the top quantized features,  which contain low-frequency motion information. It is then decoded by $DEC_{top}$ and concatenated with face features. The bottom codebook predictor then predicts the bottom codebook's logit distribution and indexes the bottom quantized features. Both quantized feature sequences are concatenated and fed to the pretrained and $DEC_{bottom}$  to produce FLAME expression $\boldsymbol{\hat{\Psi}}$ and jaw pose $\boldsymbol{\hat{\theta}^{\text{jaw}}}$ parameter sequence. During training, we utilize the straight-through Gumbel-softmax (STGS) to make indexing of codebooks differentiable, and during inference, we sample from the logit distribution by controlling each codebook's temperature $\tau_b$ and $\tau_t$.

\subsubsection{Objectives.}
We employed three distinct loss functions to optimize performance  : (i) reconstruction loss, (ii) lip loss, and (iii) emotion consistency loss.

\begin{equation}
    L_{total} = \lambda_{1}L_{rec}+\lambda_{2}L_{emo}+\lambda_{3}L_{lip}
\end{equation}

\subsubsection{Reconstruction Loss.}
We computed the Mean Squared Error between the pseudo-GT and predicted vertices, denoted as $L_{rec}$.

\subsubsection{Emotion Consistency Loss.}
To ensure that the generated face motion reflects the same emotion as the input audio, we propose an emotion consistency loss. By leveraging the trained DEE, we predict the mean from audio input $\mu_a$ and generated motion $\mu_{e}$ and enforce a high cosine similarity between these pairs to ensure they represent the same emotion. 

\begin{equation}
    L_{emo} = \frac{{\mu_{a} \cdot \mu_{e}}}{{\Vert \mu_{a} \Vert \Vert \mu_{e}\Vert}}
\end{equation}

\subsubsection{Lip Loss.}
To provide additional lip supervision, we use lip reading perceptual loss $L_{lip}$ \cite{danvevcek2023emotional}. 

\section{Experiments}
\subsubsection{Datasets.}
We utilize MEAD for train and test, and incorporate CREMA-D \cite{cao2014crema}, RAVDESS, and HDTF \cite{zhang2021flow} for evaluation. MEAD, CREMA-D, and RAVDESS are lab-recorded emotional talking face videos, while HDTF comprises YouTube-sourced talking face videos. Due to RAVDESS's limited utterances and HDTF's lack of emotion, we evaluate only emotion and lip sync, respectively. Additionally, MEAD's overlapping utterance between train and test datasets make unsuitable for fair comparisons with models untrained on it, and given no clear benchmark for lip generalization on emotional in-the-wild speeches, we constructed an audio test set, Emo-Vox, derived from VoxCeleb2. To create a reliable pseudo-3D ground truth dataset, we employ the SOTA 3D face reconstruction method \cite{danvevcek2022emoca} exclusively on lab setting datasets for accurate FLAME parameter extraction. Further details are in the Supplementary.

\subsubsection{Baseline Impelmentations.}
To compare DEEPTalk with facial parameter-based models, we train FaceFormer, FaceDiffuser, and EMOTE on MEAD dataset. As EmoTalk employs a unique training approach specifically tailored for RAVDESS and utilizes an in-house blend shape reconstruction method, we utilize its pre-trained weights. Additionally, we conduct experiments with vertex-based models (FaceFormer, FaceDiffuser, MeshTalk, CodeTalker) trained on the ground truth face mesh from the VOCASET dataset.

\subsection{Quantitative Evaluation}
\subsubsection{Evaluation Metrics.}
We adopt \textbf{FID} to evaluate the realism of rendered faces. To further assess the realism of facial movements in the FLAME parameter space, we developed \textbf{FFD (Frechet Face Distance)}, inspired by FGD \cite{FGD}, and computed it on sequences of FLAME parameters using an encoder trained on MEAD and BEATv2 \cite{EMAGE}. To evaluate emotional expressiveness, we compute \textbf{Emo-FID}, an adaptation of FID that uses an emotion feature extractor from AffectNet \cite{mollahosseini2017affectnet}, replacing the inception network to focus on emotion properties. For lip-sync evaluation, we used SyncNet metrics \cite{Chung2016OutOT} \textbf{LSE-D} (Lip Sync Error Distance) and \textbf{LSE-C} (Lip Sync Error Confidence) following \cite{aneja2023facetalk}. Unlike Lip Vertex Error (LVE), which is unsuitable for DEEPTalk due to its diverse emotional faces, these metrics evaluate lip-sync without requiring ground truth lip movements.
Finally, we measure \textbf{Diversity} \cite{diversitymetric}.

\subsubsection{Evaluation on Realism and Emotion.}
For realism comparison, DEEPTalk outperforms all other methods in FID and FFD across all datasets, shown in Table \ref{tab:combined_results}, indicating superior natural expressions and facial movements. DEEPTalk significantly surpasses other methods in Emo-FID, demonstrating superior emotional expressiveness that closely resembles real human expressions. It is noteworthy that, despite utilizing ground truth emotion labels to generate expressions, EMOTE still lags behind our approach due to its lack of diversity and tendency for exaggerated expressions.

\begingroup
\begin{table}[h]
\centering
\fontsize{9}{9}\selectfont
\begin{tabular}{lccc}
    \toprule
    Method &Train Dataset & LSE-D$\downarrow$ & LSE-C$\uparrow$ \\
    \midrule
    FaceFormer& VOCASET & 11.55 & 0.763 \\
    MeshTalk& VOCASET & 11.54 & 0.555 \\
    CodeTalker& VOCASET & 11.51 & 0.782 \\
    FaceDiffuser& VOCASET & 11.82 & 0.707 \\
    EmoTalk& RAV/HDTF & 11.40 & 0.658 \\
    FaceDiffuser& MEAD & 11.61 & 0.569 \\
    FaceFormer& MEAD & \underline{11.31} & \textbf{0.893} \\
    EMOTE& MEAD & 11.40 & 0.879 \\
    Ours& MEAD & \textbf{11.28} & \underline{0.889} \\
    \bottomrule
\end{tabular}
\caption{Lip sync results on Emo-Vox.}
\label{lip_on_vox}
\end{table}

\begin{figure*}[ht]
    \centering
    \includegraphics[width=0.99\linewidth]{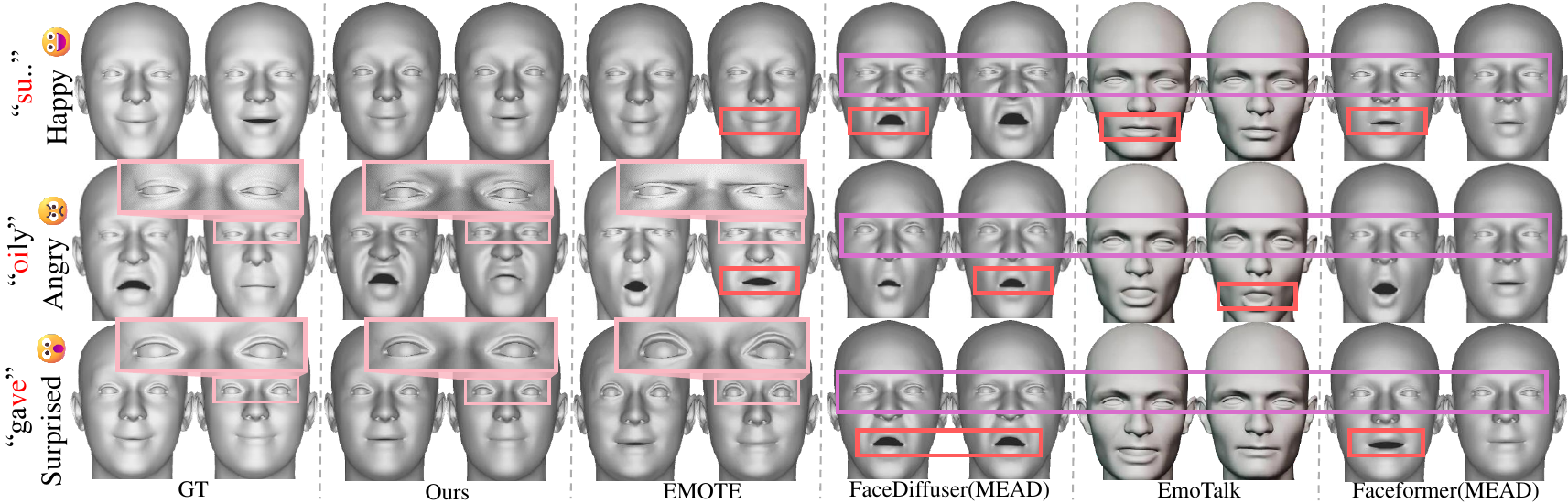}
    \caption{Qualitative results on MEAD test set. Each row displays the predicted facial motions for each utterance and corresponding emotion (left) generated by baseline models. Lip motion deviations from the ground truth are highlighted in red, while incorrect or neutral emotional expressions are indicated in purple. EMOTE, being conditioned on emotion labels, exhibits a high degree of emotional expressiveness. However, this conditioning sometimes results in exaggerated expressions, highlighted in pink in the enlarged images. In contrast, DEEPTalk generates natural emotional faces while maintaining accurate lip sync.}
    \label{fig:qualitative_baselines}
\end{figure*}

\subsubsection{Evaluation on Lip sync.}
In Table 1, DEEPTalk achieves the highest performance on LSE-D and ranks first or second on LSE-C across the MEAD test set and other datasets, demonstrating strong generalization and precise lip sync. We also conducted evaluations on Emo-Vox to assess lip sync, while ensuring a fair comparison with both parameter-based and vertex-based models.
As shown in Table \ref{lip_on_vox}, our model excels in LSE-D and ranks second in LSE-C, demonstrating effective generalization of lip movements to emotional, in-the-wild audio.
This performance further indicates our bottom codebook's capability to capture high-frequency details, enhancing dynamic lip movements.
Moreover, due to the limited scale of the ground truth scan dataset, vertex-based models fail to achieve accurate lip sync compared to those using pseudo ground truth.
Notably, FaceFormer, trained on MEAD, produces nearly neutral expressions (see Figure 3), whereas ours are more expressive, benefiting FaceFormer's lip-sync performance due to the emotion and lip-sync trade off, as demonstrated in ablation studies. 

\begingroup
\begin{table}[t]
\centering
\setlength{\tabcolsep}{1mm}
\fontsize{9}{9}\selectfont
\begin{tabular}{lcccc}
    \toprule
    Method & $\alpha$ & $\tau$ & Diversity$\uparrow$ & LSE-D$\downarrow$ \\
    \midrule
    FaceDiffuser & - & - & 21.56 & 10.93 \\
    \midrule
    Ours & 1 & argmax & 19.23 & 10.65 \\
    Ours & 0.1 & argmax & 25.48 & 10.60 \\
    Ours & -2 & argmax & 25.50 & 10.60 \\
    Ours & -4 & argmax & \textbf{25.55} & 10.60 \\
    \midrule
    Ours & mean & $\tau_t=1, \tau_b=0.1$ & 9.91 & 10.65 \\
    Ours & mean & $\tau_t = 0.1, \tau_b=1$ & 13.03 & 10.65 \\
    Ours & mean & $\tau_t = 4.5, \tau_b=1$ & \textbf{23.22} & 10.68 \\
    \bottomrule
\end{tabular}
\caption{Diversity results on the MEAD test set. DEEPTalk generates diverse faces while maintaining accurate lip sync. LSE-C reported in the Supplementary.}
\label{div}
\end{table}
\endgroup

\subsubsection{Evaluation on Diversity.}
DEEPTalk provides control over the diversity of generated facial motions in two ways: 1) \textbf{diverse emotional facial motions from the same speech input} by adjusting the control factor $\alpha$, and 2) \textbf{diverse facial motions from the same speech and emotion} by modifying the temperature of the bottom ($\tau_b$) and top ($\tau_t$) codebook predictors. We evaluated each controllability factor by varying them independently while keeping the other factor deterministic. We continued this process until our method surpassed FaceDiffuser in terms of diversity, after which we assessed lip synchronization. As shown in Table \ref{div}, DEEPTalk demonstrates superior lip synchronization, even with increased diversity, compared to FaceDiffuser on both control factors, illustrating the model’s effectiveness in generating diverse yet accurate facial motion. Notably, these two methods of control are orthogonal and can be used independently, potentially leading to even greater diversity.

\subsection{Qualitative Evaluation}

\begin{figure}[t]
    \centering
    \includegraphics[width=0.99\linewidth]{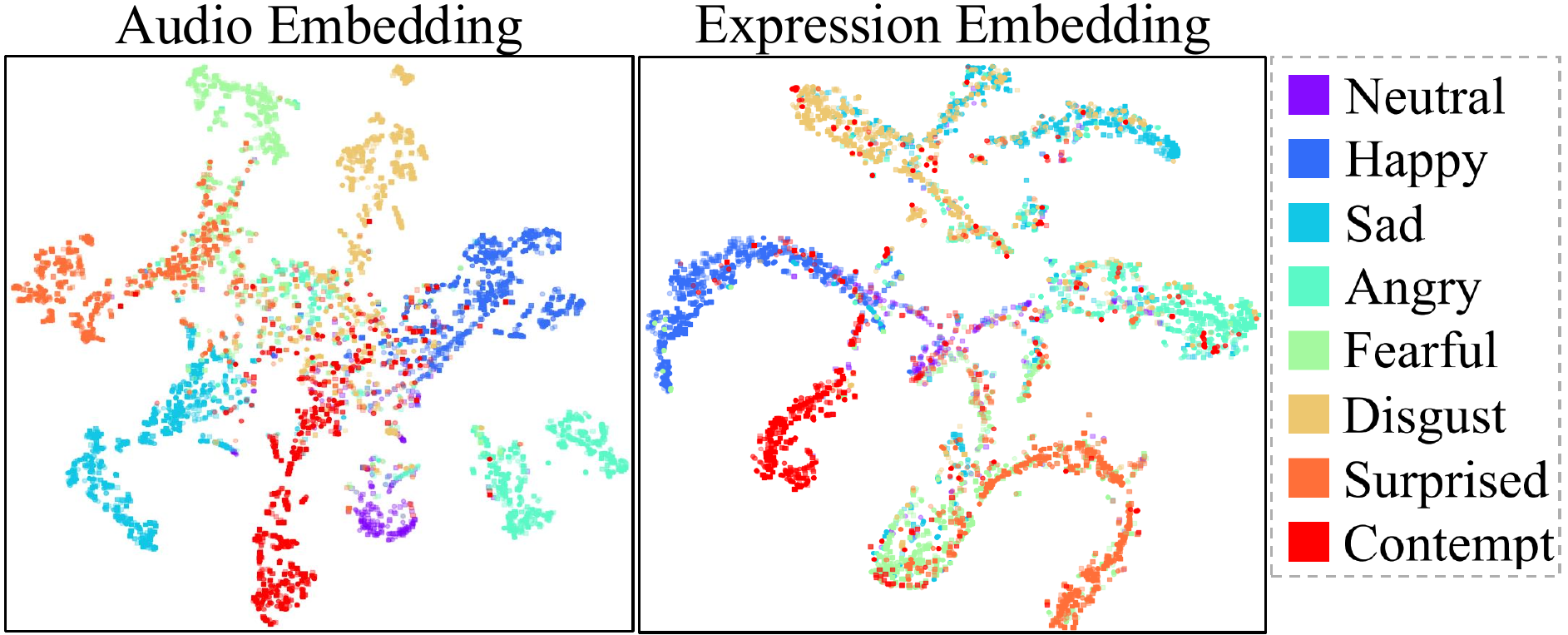}
\caption{Embeddings are clustered by emotion categories.}
    \label{fig:DEE_TSNE}
\end{figure}

\begin{figure*}[ht]
    \centering
    \includegraphics[width=0.99\linewidth]{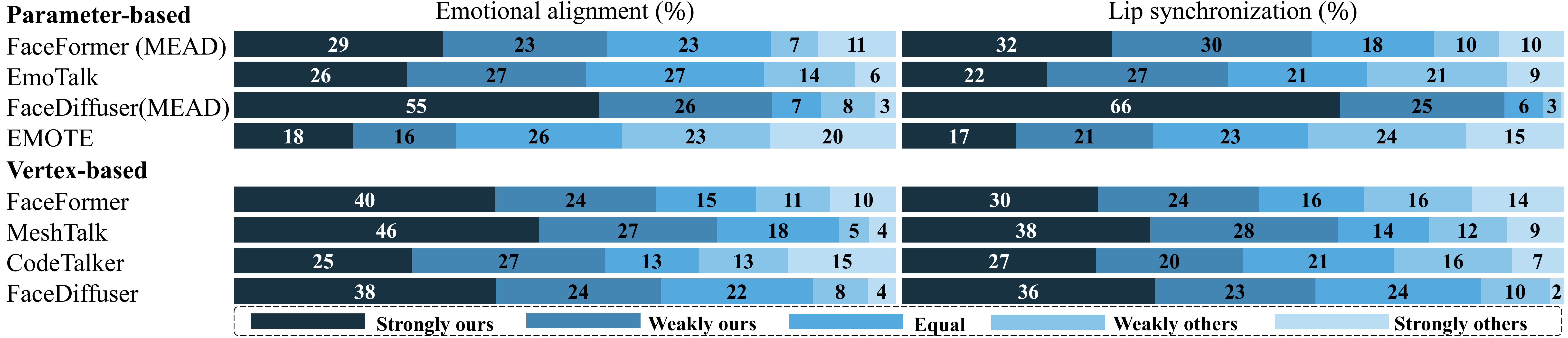}
    \caption{User Study Results. Our method is preferred over most methods on emotional alignment and lip synchronization. } 
    \label{fig:user_study}
\end{figure*}

\subsubsection{Visual Comparison.}
Figure~\ref{fig:qualitative_baselines} compares our method with SOTA methods on the MEAD test set. While most methods generate natural lip movements, they often misalign with the ground truth, such as opening or closing lips incorrectly. Additionally, except for EMOTE, other methods produce incorrect or unexpressive expressions. This highlights the effect of DEE and $L_{emo}$ on our framework. While EMOTE generates emotional expressions using emotional labels, it often produces exaggerated faces, like unnaturally wide eyes or flat eyebrows, deviating from the ground truth. In contrast, DEEPTalk generates natural emotional expressions that closely resemble the ground truth without emotion labels by leveraging emotions inherent within the speech.

\subsubsection{Effect of Emotion Embedding.} 
Figure~\ref{fig:DEE_TSNE} shows that our emotion embedding clusters by emotions using T-SNE, capturing a meaningful emotion space. Furthermore, to evaluate its efficacy in generating emotional expressions, we randomly selected two audio samples with distinct emotions and interchanged their emotion embeddings from DEE's audio encoder, before inputting them into the Face Feature Extractor. As shown in Figure~\ref{fig:emotion_swap}, this swap effectively modifies the emotional expression to match the reference speech, while maintaining precise lip synchronization with the original speech. Further analysis is provided in Supplementary.

\begin{figure}[h]
    \centering
    \includegraphics[width=0.99\linewidth]{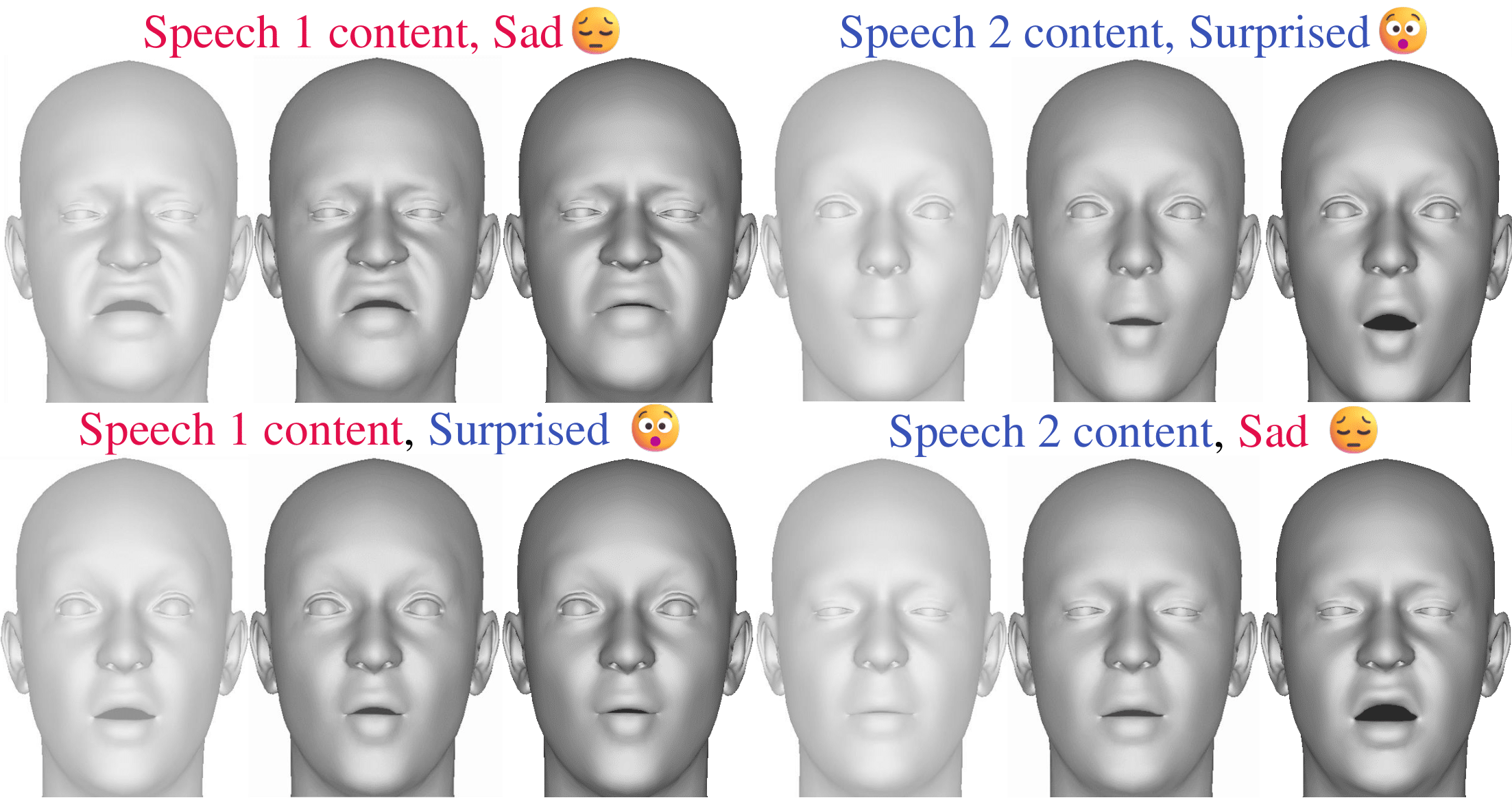}
    \caption{Swapping emotion embedding alters the emotional expression while maintaining precise lip synchronization}
    \label{fig:emotion_swap}
\end{figure}

\subsubsection{User Studies.}
We conducted A/B test with 5-point Likert scale for 98 users to evaluate model preference across two subtasks: 1) \textbf{emotional alignment} between speech and expression 2) \textbf{lip synchronization}. We sampled twenty audio clips from MEAD and ten from Emo-Vox. Details are provided in Supplementary. As shown in Figure~\ref{fig:user_study}, DEEPTalk outperformed all parameter-based models across all tasks except for EMOTE, which leverages ground truth emotion labels to generate faces, serving as the upper bound for emotional face generation. Note that EMOTE is deterministic and lacks expression diversity. For vertex-based models, DEEPTalk excelled in both tasks for all competitors. Due to the scarcity and limited size of emotional ground truth face scan datasets, vertex-based methods struggle with emotional expression and accurate lip synchronization.
This highlights that using pseudo-3D data, as done in DEEPTalk, is a promising approach for achieving accurate emotional 3D talking face animation with precise lip synchronization.

\begingroup
\setlength{\tabcolsep}{0.1mm}
\begin{table}
  \centering
  \fontsize{9}{9}\selectfont
  \begin{tabular}{lccccccc}
    \toprule
    & \multicolumn{2}{c}{\textbf{Emo-Vox}} & \multicolumn{5}{c}{\textbf{MEAD}} \\
    \cmidrule(lr){2-3} \cmidrule(lr){4-8}
    \textbf{Methods} & LSE-D$\downarrow$ & LSE-C$\uparrow$ & LSE-D$\downarrow$ & LSE-C$\uparrow$& Emo-FID$\downarrow$ & FID$\downarrow$ & FFD$\downarrow$ \\
    \midrule
    w/ VAE & 10.92 & 0.868 & 10.78 & 0.828 & 102.2 & 43.40 & 325.1\\
    w/ VQVAE & 10.87 & 1.102 & 10.70 & 1.075 & \underline{16.58} & 30.10 & 66.98\\
    w/o $L_{lip}$ & 10.98 & 1.000 & 10.91 & 0.901 & 19.94 & \underline{28.23} & \textbf{63.51}\\
    w/o $L_{emo}$ & \textbf{10.67} & \textbf{1.187} & \textbf{10.66} & \textbf{1.107} & 28.78 & 32.47 & 91.35 \\
    Full (Ours) & \underline{10.74} & \underline{1.140} & \underline{10.65} & \underline{1.103} & \textbf{15.15} & \textbf{26.07} &\underline{ 64.02} \\
    \bottomrule
  \end{tabular}
  \caption{Ablation results on Emo-Vox and MEAD test set. }
  \label{tab:freq}
\end{table}
\endgroup

\subsection{Ablation Studies}
We conducted ablation studies on the MEAD test set and Emo-Vox to analyze the contributions of individual components of DEEPTalk. Specifically, we compared (1) DEEPTalk, (2) DEEPTalk with VAE, (3) DEEPTalk with VQ-VAE, (4) DEEPTalk without $L_{lip}$ and (5) DEEPTalk without $L_{emo}$. As shown in Table \ref{tab:freq}, utilizing a discrete motion prior is crucial, as its absence results in significant quality degradation. This is due to perceptual losses like $L_{emo}$ and $L_{lip}$ where enforcing such constraints leads to unnatural expressions. Furthermore, employing TH-VQVAE enhances performance across all metrics by effectively capturing both low and high-frequency motion patterns. The incorporation of our proposed \(L_{emo}\) significantly improves emotional expressiveness and realism, as evidenced by Emo-FID, FID, and FFD scores. Interestingly, this enhancement results in a minor decrease in lip synchronization, indicating a tradeoff between emotional expressiveness and lip sync precision.

\section{Conclusion}
This paper presents DEEPTalk, a novel speech-driven talking head framework designed to generate diverse and emotionally expressive facial animations. Unlike previous methods, DEEPTalk achieves precise lip synchronization while ensuring facial expressions accurately reflecting the emotional tone of the input speech. This is accomplished through our dynamic emotion embedding (DEE), which serves as a strong emotion prior, and a temporally hierarchical VQ-VAE (TH-VQVAE), which functions as a robust and dynamic motion prior. Additionally, the probabilistic design of DEEPTalk facilitates non-deterministic generation that can be controlled on various levels. Extensive experiments on various datasets demonstrate the superiority of our model over existing methods across six metrics, with notable improvements in emotional realism.

\section{Acknowledgments}
This work was supported by an IITP grant funded by the Korean Government (MSIT) (No. RS-2020-II201361 , Artificial Intelligence Graduate School Program (Yonsei University)) and Culture, Sports and Tourism R\&D Program through the Korea Creative Content Agency grant funded by the Ministry of Culture, Sports and Tourism in 2024 (Project Name:Development of multimodal UX evaluation platform technology for XR spatial responsive content optimization, Project Number: RS-2024-00361757) and GIANTSTEP Inc.
\bibliography{aaai25}

\clearpage
\setcounter{page}{1}
\renewcommand{\thesection}{\Alph{section}}
\setcounter{section}{0} 
\renewcommand{\thetable}{\Alph{table}}
\setcounter{table}{0}
\renewcommand{\thefigure}{\Alph{figure}}
\setcounter{figure}{0}

\definecolor{codebg}{rgb}{0.95, 0.95, 0.95}  
\lstdefinestyle{mystyle}{
    backgroundcolor=\color{codebg},   
    basicstyle=\ttfamily\small,       
    frame=single,                     
    breaklines=true,                   
    captionpos=b,                      
    keywordstyle=\bfseries,            
    commentstyle=\color{gray},         
    numbersep=5pt,                     
    xleftmargin=5pt, xrightmargin=5pt  
}

\twocolumn[
    \centering
    \LARGE 
    \textbf{DEEPTalk: Dynamic Emotion Embedding for Probabilistic Speech-Driven 3D Face Animation}\\
    \vspace{0.5em}Supplementary Material \\
    \vspace{2em}
]

\begin{abstract}
This supplementary material includes the following sections: 
(1) Implementation details of our proposed models, 
(2) Extended analysis of our Dynamic Emotion Embedding, 
(3) Comparison of 3D Talking Head Datasets, 
(4) Details of our evaluation metrics, and 
(5) Additional qualitative results.

\end{abstract}

\section{Method Details}
This section provides detailed descriptions of the DEE, TH-VQVAE, and DEEPTalk models, including specifics on their architectural design and training details.

\begin{figure}[h!]
    \centering
    \includegraphics[width=0.8\linewidth]{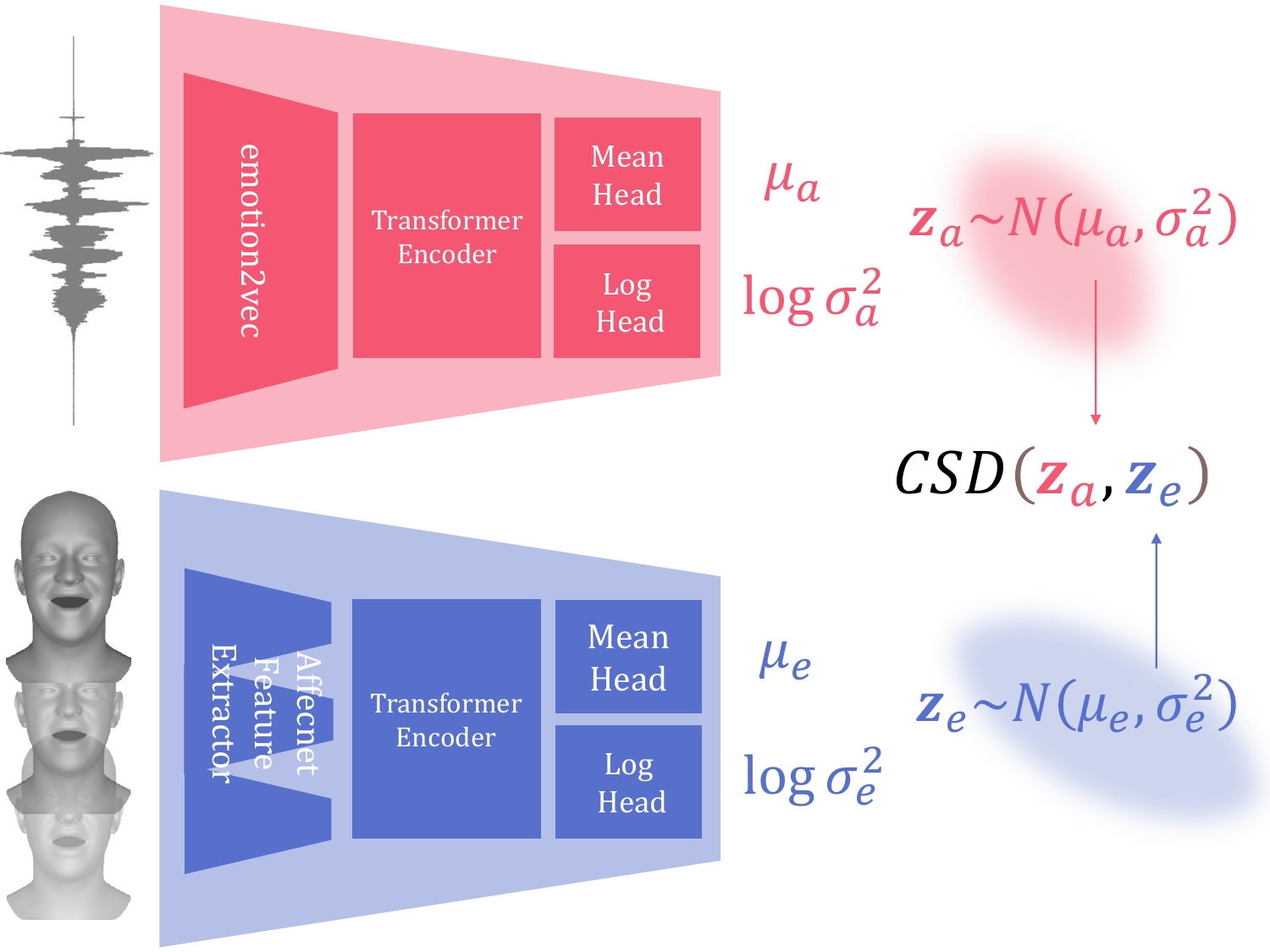}
   \vspace{-5pt}
    \caption{Detailed architecture of DEE}
   \vspace{-5pt}
    \label{fig:DEE architecture}
\end{figure}

\subsection{DEE}
DEE comprises two feature extractors and two encoders for both audio and expression modalities. In the audio feature extractor, we employed a pretrained emotion2vec model \cite{ma2023emotion2vec}, with six unfrozen layers from the final layer and the remaining layers frozen. As for the expression feature extractor, a simple Multi-Layer Perceptron (MLP) was utilized, featuring hidden layer sizes of 256, 512, 1024, 512, and 128, with LeakyReLU activations. This MLP was trained on the Affectnet dataset, but with FLAME parameters rather than images, to classify 8 emotions. We obtain an accuracy of 56.64 on the Affectnet benchmark being close to 58 on the original Affectnet paper \cite{mollahosseini2017affectnet}. 

For the audio encoder, a two-layer transformer with eight heads and a model dimension of 128 was employed, while the expression encoder utilized a six-layer transformer architecture. Each mean head and log head comprised a single-layer transformer encoder and a generalized pooling operator(GPO)\cite{GPO} layer. Notably, the output of the mean head was normalized, whereas the log head output remained unnormalized. Detailed model architecture is shown in Figure \ref{fig:DEE architecture}.

The model underwent training with a learning rate set to 0.0001 for 300 epochs, utilizing a cosine annealing scheduler. The FLAME parameters had a length of 50, while audio samples were set to 32000, resulting in 2-second sliced clips. Clips longer than 2 seconds were randomly sliced, while shorter clips were padded accordingly.
\begin{figure}[h]
    \centering
    \includegraphics[width=0.9\linewidth]{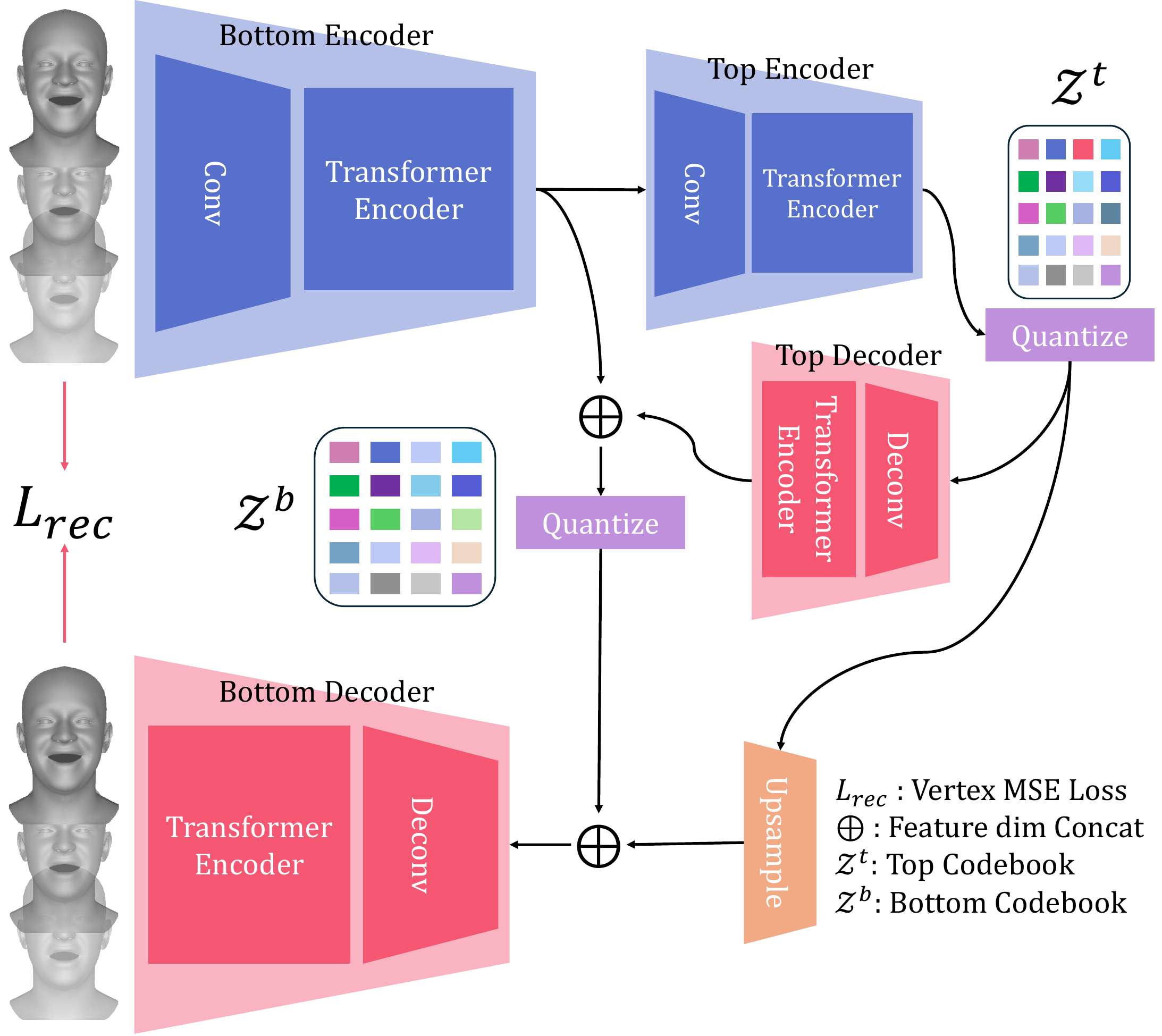}
   \vspace{-5pt}
    \caption{Detailed architecture of TH-VQVAE}
   \vspace{-5pt}
    \label{fig:TH-VQVAE_architecture}
\end{figure}

\begin{figure*}[h]
    \centering
    \includegraphics[width=0.99\linewidth]{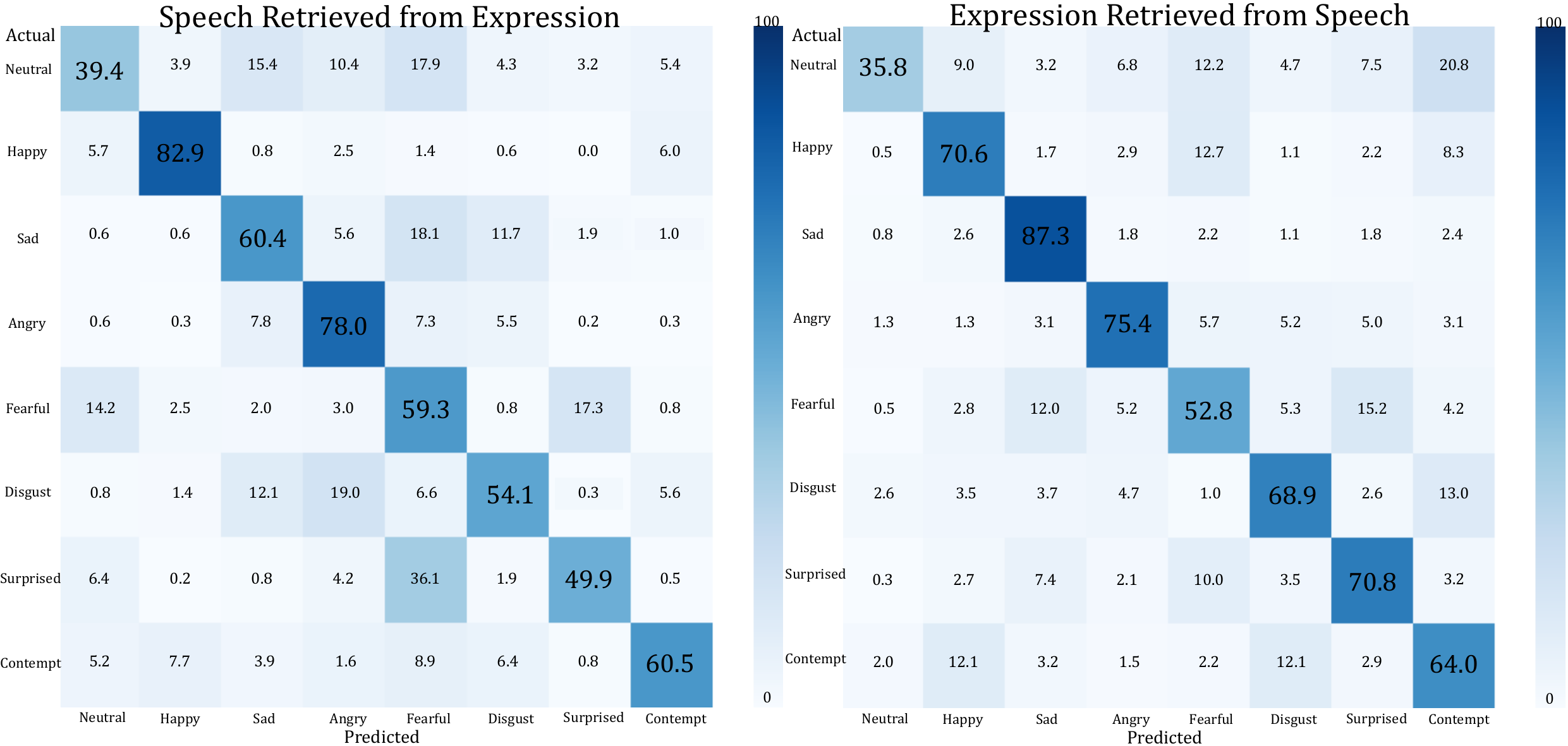}
\caption{Inter-modal Retrieval Top-1 Accuracy. Our model effectively retrieves corresponding speech or facial expressions sharing the same emotion. Neutral emotions present a greater challenge for retrieval compared to other emotions due to the reduced information available in both speech and facial expressions.}
    \label{fig: Inter-modal retreiveal}
\end{figure*}

\begin{figure*}
    \centering
    \includegraphics[width=0.9\linewidth]{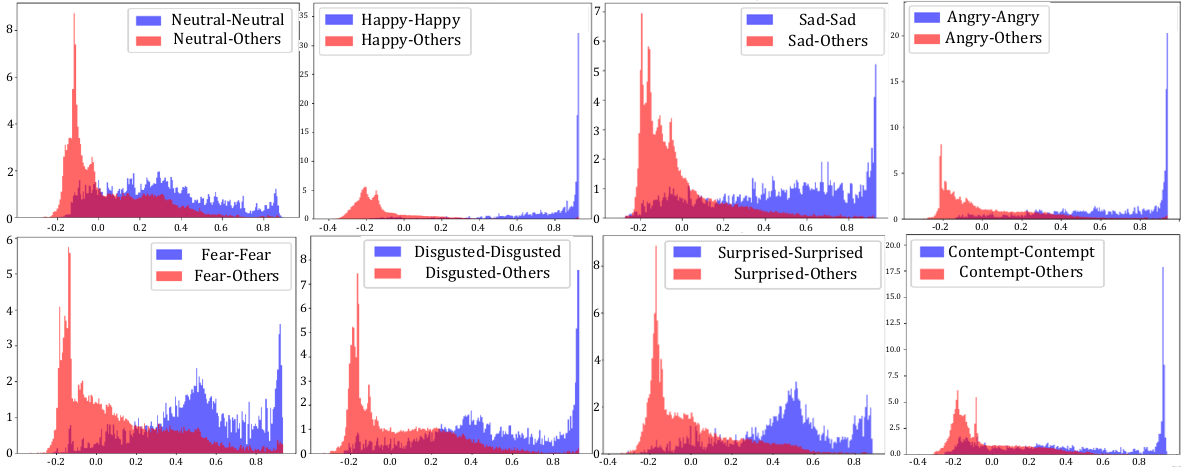}
    \caption{Histogram of cosine similarities for all inter-modal pairs. Embeddings with identical emotional labels exhibit an elevated similarity, denoted by blue coloration, whereas embeddings representing different emotional states demonstrate a diminished level of similarity, depicted in red.}

    \label{fig:cossim}
\end{figure*}

\subsection{TH-VQVAE}
The Temporally Hierarchical VQ-VAE architecture consists of two encoders and two decoders, both top and bottom. Each encoder comprises a convolutional layer with a quantization length of 1, succeeded by a single-layer transformer. Likewise, each decoder includes a single-layer deconvolutional layer responsible for decoding latent embeddings along the temporal axis, followed by a single-layer transformer. The top codebook consists of 256 codes, each with a dimensionality of 128, while the bottom codebook comprises 256 codes, each with a dimensionality of 256. Detailed model architecture is shown in Figure \ref{fig:TH-VQVAE_architecture}. 

The temporally hierarchical VQ-VAE (TH-VQVAE) model was trained with a learning rate of 0.005 over 1500 epochs, utilizing a batch size of 256. A learning rate warm-up strategy was employed for the initial 4000 steps of training and the reconstruction loss weight was set to 100,000, while the quantization loss weight was set to 1. As the reconstructed FLAME parameters were noisy, we followed \cite{learning2listen_motionprior} using the savgol filter to smooth movements with respect to the temporal axis. 

\subsubsection{Loss Function.} We define TH-VQVAE's reconstruction loss as the $L_2$ distance between the reconstructed vertex and the input vertex :
$$
L_{\text {rec }}=\|\widehat{\mathbf{V}}_{1: T} -  \mathbf{V}_{1: T}\|_2^2,
$$
where $\mathbf{V}_t,\widehat{\mathbf{V}}_t$ are obtained by $\mathbf{V}^t=\text{FLAME}\left( \boldsymbol{\psi}_t, \boldsymbol{\theta}^{\text {jaw }}_t\right)$. 
The final loss function for training TH-VQVAE is as following:
\begin{equation}
\mathcal{L}_\text{total}= L_{\text {rec }} +\beta\|Z-\operatorname{sg}[C]\|,
\end{equation}
where, $sg[\cdot]$ signifies the stop gradient operation, and $\beta$ denotes the weight of the "commitment loss." The updating of the codebook is facilitated through the use of an Exponential Moving Average (EMA)\cite{VQVAE2}.

\subsection{DEEPTalk}
The DEEPTalk talking head generator consists of two main components: the Face Feature Extractor and the Codebook Predictors (top and bottom). The Face Feature Extractor includes an audio encoder and a transformer encoder. For the audio encoder, we utilize a pre-trained Wav2vec2 model \cite{baevski2020wav2vec}, where the Temporal Convolutional Network (TCN) layers are frozen, and the remaining layers are unfrozen. The embeddings from this audio encoder are concatenated with style embeddings, which are derived by concatenating the DEE’s audio encoder output with an identity one-hot vector, followed by a linear layer. These concatenated features are then input to a single-layer transformer encoder with 128 model dimensions and 8 heads.

For the Codebook Predictors, both the top and bottom predictors are 6-layer transformers with 8 heads and a 128-dimensional feature space. Using the face features, the top codebook predictor first predicts the top codebook index, and the resulting top feature is decoded by the top decoder of TH-VQVAE and then concatenated with the output of the Face Feature Extractor. This concatenated feature is then passed to the bottom codebook predictor to predict the logits of the bottom codebook. The final output is obtained by upsampling the top quantized feature and concatenating it with the bottom quantized feature. This output is fed into the TH-VQVAE bottom decoder to generate the final FLAME parameters.

The training process involved two stages. In the first stage, only the vertex error was utilized for training. In the second stage, an emotion consistency loss and a lip reading loss were introduced. Each stage was trained separately with a learning rate of 0.0001, a batch size of 128 for the first stage, and 16 for the second stage. The first stage was trained for 200 epochs, while the second stage was trained for 1 epoch. In the first stage, the tau of each gumbel softmax was set to decrease linearly from 2 to 0.1, while in the second stage, we used 2.


\begin{table*}[h]
\centering
\scriptsize 
\begin{tabularx}{\textwidth}{@{}l l l l l l l r@{}}
\toprule
\textbf{Dataset} & 
\parbox{1.8cm}{\centering \textbf{Emotional\\Talking\\Duration}} & 
\parbox{1.8cm}{\centering \textbf{Representation}} & 
\parbox{1.8cm}{\centering \textbf{Capture\\Method}} & 
\parbox{1.8cm}{\centering \textbf{Scalability(reason)}} & 
\parbox{1.8cm}{\centering \textbf{Emotion\\Intensity}} & 
\parbox{1.8cm}{\centering \textbf{Emotionally\\Acted Expressions}} & 
\textbf{\#Actors} \\ \midrule
BIWI & 0h & GT vertex & 3D Scan & $\times$ (Scan) & $\times$ & $\times$ & 14 \\
VOCASET & 0h & GT vertex & 3D Scan & $\times$ (Scan) & $\times$ & $\times$ & 12 \\
BEAT & 33.19h & MC-BS & iPhone & $\times$ (Depth Info) & $\times$ & $\times$ (gestures) & 30 \\
BEATv2 & 33.19h & MC-FLAME  & iPhone & $\times$ (Depth Info) & $\times$ & $\times$ (gestures) & 30 \\
3D-ETF & 1.5h & PGT-BS & in-house recon & $\times$ (Closed Source) & \checkmark(RAVDESS) & \checkmark(RAVDESS) & 324+ \\
3D-RAVDESS & 1.5h & PGT-FLAME & EMOCAv2 & \checkmark (Open) & \checkmark & \checkmark & 24 \\
3D-CREMA-D & 5.23h & PGT-FLAME & EMOCAv2 & \checkmark (Open) & \checkmark & \checkmark & 91 \\
3D-MEAD & \textbf{36h} & PGT-FLAME & EMOCAv2 fine-tuned on MEAD & \checkmark (Open) & \checkmark & \checkmark &60 \\ 
\bottomrule
\end{tabularx}
\caption{Comparisons of various Datasets. ‘PGT’
denotes Pseudo Ground Truth, ‘MC’ denotes Motion Capture and 'BS' denotes blendshapes. }
\label{tab:dataset_comparison}
\end{table*}
\section{DEE Analysis}
To demonstrate the comprehensive nature of our Dynamic Emotion Embedding (DEE) as an emotional embedding space that encompasses both expression and speech, we conducted two retrieval tasks and visualized histograms of cosine similarities using the MEAD test set.
    
\textbf{Inter-modal retrieval} Two inter-modal retrieval tasks were conducted: 1) retrieving speeches sharing the same emotion label as a given expression and 2) retrieving expressions sharing the same emotion label as a given speech. The top-1 accuracy of DEE's inter-modal task is depicted in Figure \ref{fig: Inter-modal retreiveal}. The capacity to retrieve speech or expression from the alternative modality underscores the establishment of an emotional joint embedding space within DEE's framework, accommodating both expressive facial features and speech signals.


\textbf{Cosine Similarity:} For each emotional label, we calculated the histogram of embeddings' cosine similarity for all inter-modal pairs with different emotions and the same emotions. As shown in Figure \ref{fig:cossim}, embeddings with identical emotional labels exhibit an elevated similarity, denoted by bluecoloration, whereas embeddings representing different emotional states demonstrate a diminished level of similarity, depicted in red. This demonstrates how the embeddings are clustered by emotions, for both speech and expression.


\section{Datasets}
\subsection{3D Talking Head Dataset}
 While 3D talking head datasets like BIWI\cite{BIWI} and VOCASET\cite{VOCA2019} provide accurate data, they lack emotional expressions and scalability, making inadequate for learning emotional facial dynamics. The BEAT dataset, introduced by \cite{BEAT}, presented a large-scale co-gesture dataset utilizing an iPhone to capture the facial expressions of speaking actors. However, as shown in Table \ref{tab:dataset_comparison}, BEAT comprises only 30 actors and lacks emotion intensities. Crucially, it is designed for co-speech gesture generation rather than talking heads, limiting the emotional expressiveness of facial animations since actors focused on gestures over facial expressions. Additionally, the use of non-personalized ARKit blendshapes captured by an iPhone compromises its quality. Recently, EMAGE \cite{EMAGE} introduced BEATv2, which maps BEAT's blendshape weights to FLAME parameters with the assistance of animators. While this conversion enhances the dataset's utility for academic research, it still falls short in capturing subtle movements and expressiveness of facial motion, as it is derived from the original BEAT blendshape weights.
On the other hand, to train an emotional talking head model, \cite{peng2023emotalk} curated a pseudo-3D dataset, called 3D-ETF composed of blendshape weights captured using an in-house reconstruction method applied to RAVDESS and HDTF. Despite its considerable scale and inclusion of over 324 actors, the dataset's emotional talking duration is limited to just 1.5 hours, identical to RAVDESS, as HDTF lacks emotional content. Furthermore, RAVDESS includes only 2 sentences, severely limiting the variety of utterances. These limitations pose significant challenges for generalizing to a broader range of emotional speeches.

 To address the limitations of existing datasets, we constructed a pseudo-3D dataset from the MEAD video dataset (3D-MEAD) to create a large-scale emotional dataset, following the approach of EMOTE \cite{danvevcek2023emotional}. The MEAD dataset is large-scale, featuring 60 actors speaking diverse sentences with varying levels of emotion. This extensive range makes it suitable for training models that generalize well to various emotional expressions. We employed a fine-tuned version of EMOCAv2 \cite{danvevcek2022emoca} on MEAD to achieve more accurate reconstructions. Moreover, the MEAD dataset comprises video recordings captured in a lab setting, focusing on the frontal view of the upper body with a fixed camera, which minimizes noise and facilitates the generation of precise 3D pseudo labels. Similarly, CREMA-D and RAVDESS share this controlled recording characteristic, making them suitable for benchmarking. In contrast, in-the-wild video datasets like HDTF and VoxCeleb contain noise from factors such as head turns, resulting in poor reconstruction outcomes. Consequently, we exclude in-the-wild datasets from the construction of our 3D pseudo-dataset, utilizing these datasets exclusively for evaluating lip synchronization, as they do not require 3D vertex data.
\subsection{Emo-Vox}
We introduce a test set for the emotional talking head task, named Emo-Vox, for two primary reasons. First, to evaluate whether models can generate synchronized lip motion with emotional, in-the-wild speeches. Second, to validate performance on previously unseen utterances, since the same utterances exist between MEAD's training and test sets.  
Therefore, we curated emotional speeches from VoxCeleb2 to create Emo-Vox. We employed the DEE encoder, utilizing high-intensity emotion embeddings from MEAD as anchors to retrieve high-intensity emotional speeches. By extracting and comparing the emotion embeddings of audio from VoxCeleb2 with high-intensity emotion embeddings from MEAD (both audio and facial motion), we selected videos with similar emotional cosine similarity. Consequently, we curated a total of 1.3 hours of in-the-wild, emotional test data featuring diverse utterances. Note that while our curation pipeline allows for significant expansion of this test set, we limited its size to be comparable to that of RAVDESS.

\section{Metrics}
\subsection{Frechet Face Distance} While the MEAD dataset contains emotional and expressive facial motions, it remains a pseudo-3D data set. Relying solely on MEAD to train an encoder for Frechet Facial Distance (FFD) may be unfair and show bias. To address this, we combine the MEAD dataset with BEATv2 to form a comprehensive dataset that includes both emotional facial expressions and ground truth motion capture data. BEATv2, which originates from the initial BEAT dataset \cite{BEAT} in blend shape weights, was later transformed into FLAME space by \cite{EMAGE}. Although BEATv2 is less expressive and less emotional compared to MEAD, it is regarded as providing ground truth motion data.

We develop a Variational Autoencoder (VAE) whose encoder comprises 1D convolutional layers to downsample FLAME motion sequences by a factor of 8, followed by a single-layer transformer. The decoder features a single-layer transformer succeeded by 1D upconvolutional layers that upsample the latent representation back to the original FLAME sequences. Utilizing the trained encoder of the VAE, we calculate the Frechet distance between the latent features of human facial motions, denoted as \( F \), and the latent features of predicted facial motions, denoted as \( \hat{F} \). The Frechet Facial Distance (FFD) is defined as:
\[
\text{FFD}(F, \hat{F}) = \| \mu - \hat{\mu} \|^2 + \text{Tr}(\Sigma + \hat{\Sigma} - 2\sqrt{\Sigma \hat{\Sigma}})
\]

Here, \( \mu \) and \( \hat{\mu} \) represent the mean vectors of the real and predicted feature distributions, respectively, while \( \Sigma \) and \( \hat{\Sigma} \) denote the corresponding covariance matrices. This metric provides a comprehensive measure of the similarity between the distributions of the real and generated facial motions.

\begin{figure}[h]
    \centering
    \includegraphics[width=0.9\linewidth]{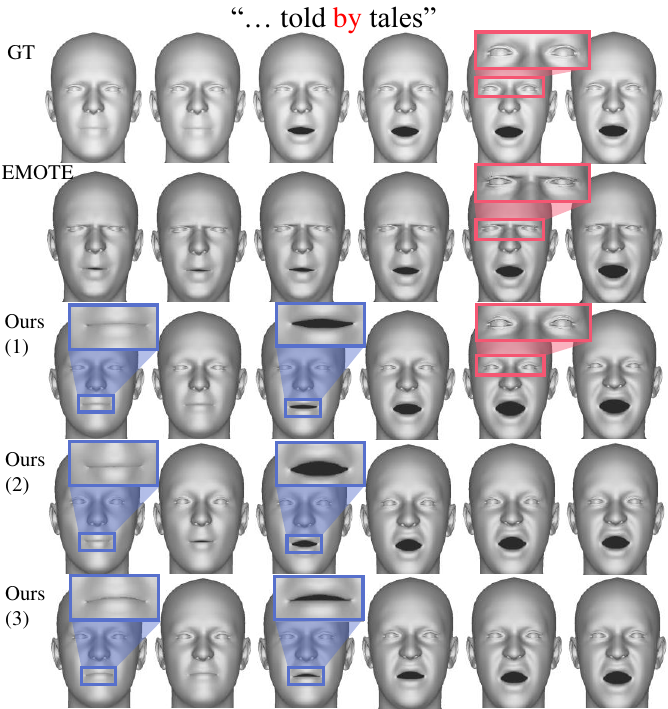}
   \vspace{-5pt}
\caption{{Diverse Emotional Expressions from the Same Audio.} The first row displays a sequence of ground truth rendered faces saying the word "by." The second row shows the prediction by EMOTE, while the subsequent rows present three individual predictions by DEEPTalk. Our method generates diverse emotional expressions (shown in blue) and produces natural emotional expressions close to the ground truth (shown in pink).
 }

   \vspace{-5pt}
    \label{fig:diverse_emotion_from_same_audio}
\end{figure}

\subsection{Lip sync evaluation}
In the case of lip synchronization, we used the lip sync metric from \cite{Chung2016OutOT}. Since our method generates diverse facial expressions from the same audio, resulting in different lip shapes for the same pronunciation (e.g., the corners of the mouth rise when smiling), it becomes challenging to directly compare the lip vertices with the ground truth (GT). Therefore, by employing the lip sync metric, we can measure how well the rendered lip-cropped image matches the input audio, not ground truth vertices. Specifically, the generated faces are rendered, cropped, and fed into a pre-trained SyncNet Image model. The MFCC spectrum of the corresponding audio is input into the pre-trained SyncNet Audio model. The pairwise distance between both sets of features is then calculated to evaluate the lip sync distance where a smaller distance means an accurate lip sync. 


\begin{figure}
    \centering
    \includegraphics[width=0.9\linewidth]{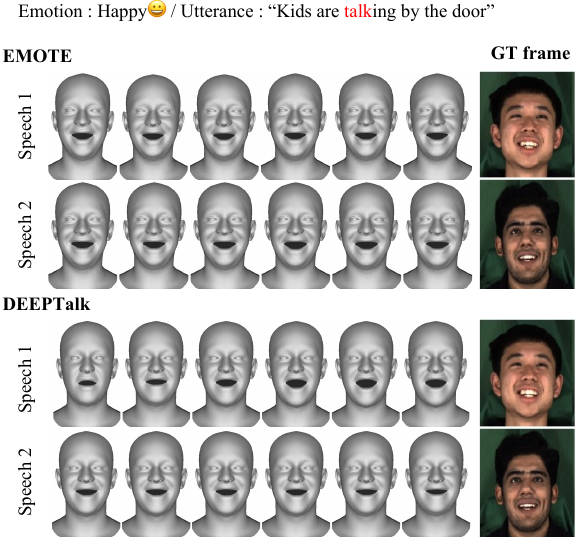}
   \vspace{-5pt}
    \caption{{Different expression from different speech with happy emotion label. } For the same emotion, EMOTE produces identical expressions, whereas DEEPTalk generates diverse expressions.}
   \vspace{-5pt}
    \label{fig:diverse_exp_happy}
\end{figure}

\begin{figure}
    \centering
    \includegraphics[width=0.9\linewidth]{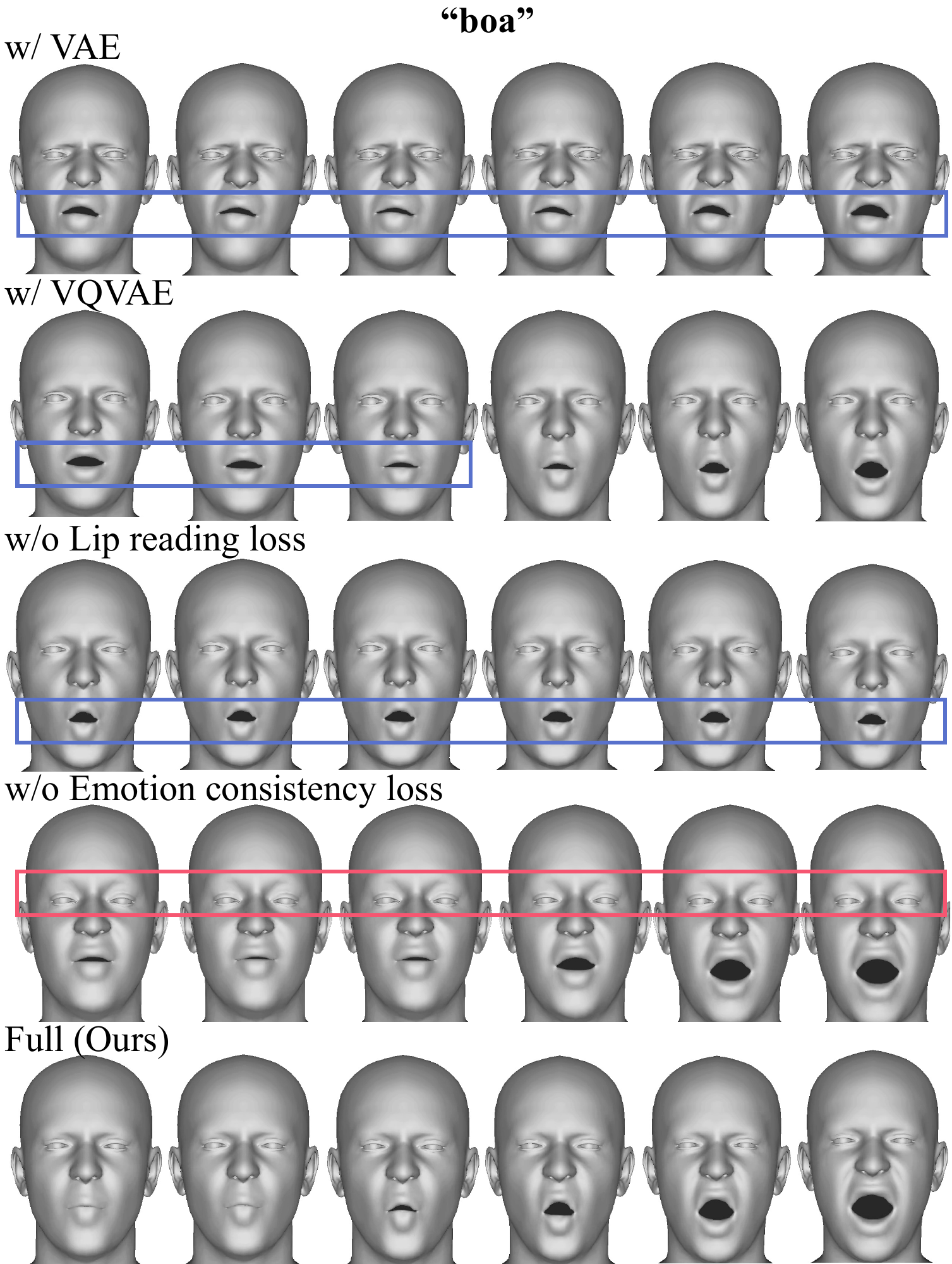}
   \vspace{-5pt}
    \caption{Ablation study. Each row illustrates the effects of different model components being replaced or removed. For the VAE, the lip shapes appear awkward and generate unnatural expressions. The VQVAE struggles to close the lips properly during rapid utterances, such as when pronouncing "b". When the lip reading loss is absent, the overall lip movements do not align well, and without the emotion consistency loss, the generated faces are unexpressive and not emotional.}
   \vspace{-5pt}
    \label{fig:ablation}
\end{figure}

\section{Additional Qualitative Results}
\subsection{Diverse emotion from the same audio input}
DEEPTalk can generate diverse emotional facial motions from the same speech input by sampling from the emotion distribution. We provide visuals of the generated facial motion in Figure 5.



\renewcommand{\arraystretch}{1.3}
\begin{table*}[ht]
    \centering
    \begin{tabular}{c|ccccc|ccccc}
    \toprule
        \centering
        \multirow{2}{*}{Ablation} &
        \multicolumn{5}{c|}{\textbf{Emotional alignment}} & \multicolumn{5}{c}{\textbf{Lip synchronization}} \\ \cline{2-11}
        
        & \vspace{0.0cm} 5 & \vspace{0.0cm} 4 & \vspace{0.0cm} 3 & \vspace{0.0cm} 2 & \vspace{0.0cm} 1 & \vspace{0.0cm} 5 & \vspace{0.0cm} 4 & \vspace{0.0cm} 3 & \vspace{0.0cm} 2 & \vspace{0.0cm} 1\\
        \midrule
        \begin{tabular}[c]{@{}c@{}}w/ VAE\end{tabular}  & \cellcolor{orange!50}0.229  & \cellcolor{red!50}\textbf{0.327} & 0.180 & 0.200 &  0.0634 & \cellcolor{orange!50}0.356 & \cellcolor{red!50}\textbf{0.360}& 0.156 & 0.0927 & 0.0341\\
        \cline{1-1}
        \begin{tabular}[c]{@{}c@{}}w/ VQVAE\end{tabular} & 0.186  &\cellcolor{orange!50}0.277 & \cellcolor{red!50}\textbf{0.394} & 0.0945 &  0.0489 & 0.205  & \cellcolor{red!50}\textbf{0.362} & \cellcolor{orange!50}0.248 & 0.111 &  0.0749\\
        \cline{1-1}
        \ w/o $L_{lip}$ & \cellcolor{red!50}\textbf{0.507}  & 0.197 & \cellcolor{orange!50}0.207 & 0.0526 &  0.0362 & \cellcolor{red!50}\textbf{0.623}  & \cellcolor{orange!50}0.238 & 0.0960 & 0.0265 &  0.0166\\
        \cline{1-1}
        \begin{tabular}[c]{@{}c@{}}w/o $L_{emo}$\end{tabular} & 0.197 & \cellcolor{red!50}\textbf{0.278} & \cellcolor{orange!50}0.261 & 0.146 & 0.119 & 0.115 & \cellcolor{orange!50}0.224 & \cellcolor{red!50}\textbf{0.322} & 0.214 & 0.125 \\
    \bottomrule
    \end{tabular}
\caption{User study results for ablation. 5 indicates DEEPTalk strongly preferred and 1 indicates others strongly preferred. The highest score is denoted by a red box, and the second highest score is represented by an orange box.}
\label{tab:userstudy_ablation}
\end{table*}

\subsection{Diverse expression from different speech with the same emotion label}
Emotional datasets and models that utilize emotion labels as input typically categorize emotions into 7 or 8 classes. However, even within the same emotion category, facial expressions can vary significantly as there are diverse ways of expressing an emotion. Models like EMOTE fail to capture this variation, generating identical expressions for the same emotion label even when the speech differs. In contrast, DEEPTalk can generate diverse expressions for audio that have the same emotion label. Figure 6 presents the generated results using the RAVDESS dataset, where speech1 and speech2 refer to the speeches of two different actors acting the same sentence with the same emotion. In the GT frames of Figure 6, it can be observed that even when expressing the same emotion, each person's facial expression differs. In contrast, EMOTE produces the same expressions for different speeches. However, DEEPTalk generates emotional expressions that align with the emotion label but with varying styles, resembling the natural variations observed among different individuals.

\subsection{Ablation study}
We conducted an ablation qualitative test to evaluate how each component of our model influences the generated facial expressions, as shown in Figure 7. The test was performed on the Emo-Vox dataset.


\subsection{User Studies}

Extensive user studies were conducted for parameter-based methods, vertex-based methods, and ablations. Before executing the tasks, as in Figure \ref{fig:userinfo for user study}, participants' information was first surveyed to eliminate potential biases and ascertain the relevance of the participants with our survey. Subsequently, as shown in Figure~\ref{fig:instruction for user study} , comprehensive instructions were provided, and as shown in Figure~\ref{fig:confirm for user study}, confirmatory questions were used to ensure that participants understood the instructions well. Responses from participants who demonstrated misunderstanding during this phase were disregarded.
We follow the A/B user study methods done in recent works \cite{stan2023facediffuser} and use a 5-point Likert scale\cite{danvevcek2023emotional}. As using the whole test set is not feasible, we randomly sample four and four audio samples from MEAD and emo-voxceleb and use them as our test case, resulting in a total of 98 participants across all three tasks. 

\textbf{Ablation study}
Additionally, we included a user study table specifically for the ablation, which was not featured in the main paper, to provide further insights into our analysis. This ablation study evaluates the influence of individual components of the DEEPTalk model on perceptual quality. As in Table \ref{tab:userstudy_ablation}, a rating scale ranging from 1 to 5 was employed, where 5 indicates our DEEPTalk is strongly better, and 1 indicates another baseline is strongly better. The highest score is denoted by a red box, whereas the second-highest score is represented by an orange box. The ablation components examined in this study mirror those of the quantitative ablation study in the main paper. Given pairs of rendered videos, participants were then prompted to select a better method, considering each aspect, like the other experiments.

For relevancy of emotion for audio and generated facial motion sequence, Table~\ref{tab:userstudy_ablation} demonstrates that our approach achieves the highest performance when each of the four components is eliminated. Moreover, for the model without emotion consistency loss, ours shows further improvements which indicate that our perceptual emotion consistency loss, utilizing DEE, helps the model to generate more audio aligned emotion facial motion. Furthermore, for models without discrete motion prior and hierarchical codebook, our approach achieves a high score due to TH-VQVAE's reconstruction capabilities. This allows DEEPTalk to generate a more expressive facial motion with the same emotion supervision. Discretized motion prior solves the smoothing problem, while the hierarchical codebook learns to reconstruct dynamic frequency signals effectively, thus leading to generating more expressive facial motion.

Regarding lip synchronization, DEEPTalk exhibits significantly better performance with lip reading loss, as indicated in Table~\ref{tab:userstudy_ablation}. This suggests that perceptual lip loss supervision effectively guides DEEPTalk to generate realistic lip movements. Furthermore, our method achieves a significantly higher score compared to approaches without discrete motion prior and without a hierarchical codebook. This is due to TH-VQVAE's ability of generating high-frequency motion, such as lip movements. Additionally, for emotion consistency loss, our result is comparable to the model without emotion consistency loss, indicating that our emotion consistency loss does not deteriorate lip movements' quality.



\begin{figure*}[h]
    \centering
    \includegraphics[width=0.9\linewidth]{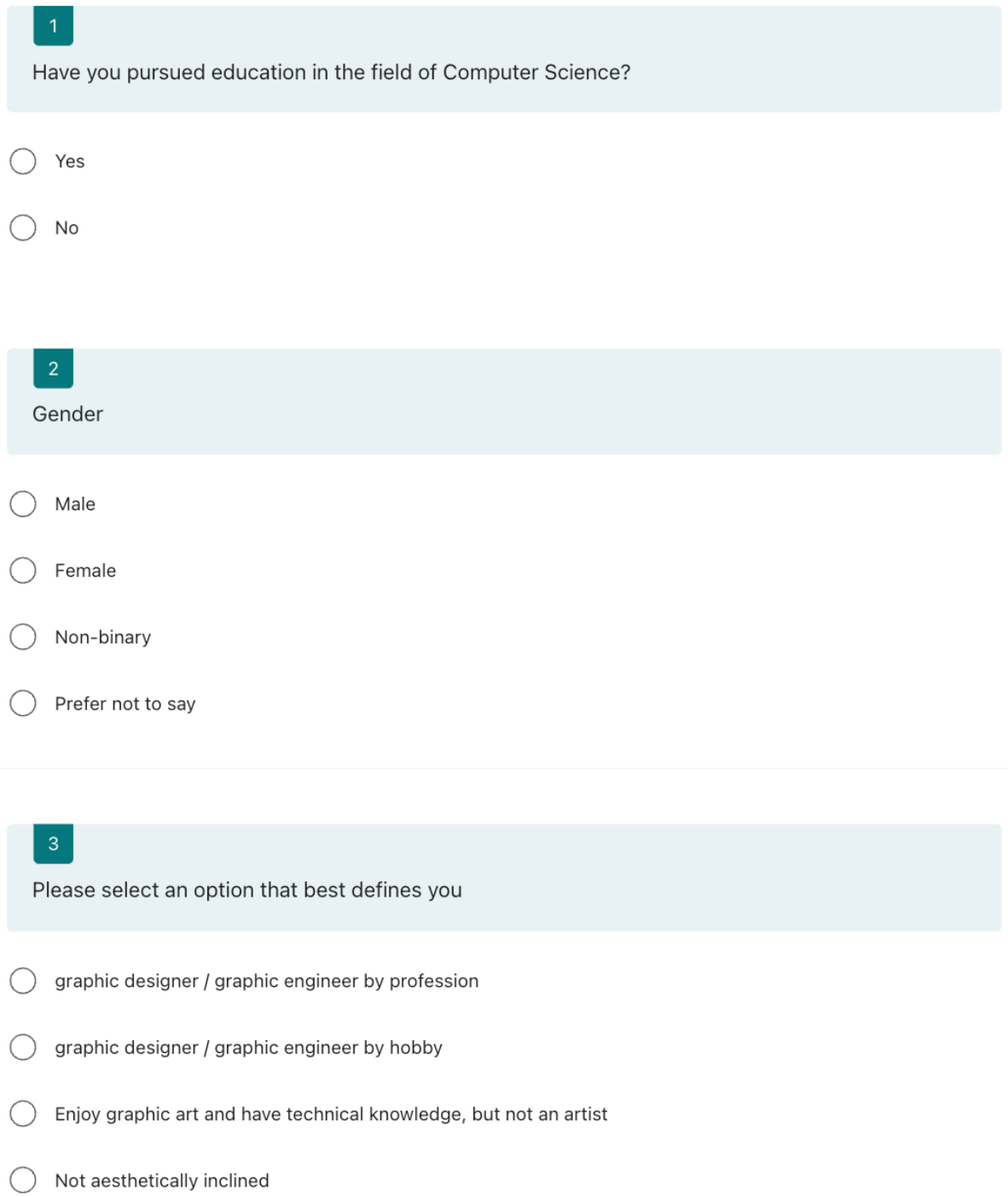}
   \vspace{-5pt}
    \caption{User info for User Study}
   \vspace{-5pt}
    \label{fig:userinfo for user study}
\end{figure*}

\begin{figure*}[h]
    \centering
    \includegraphics[width=0.9\linewidth]{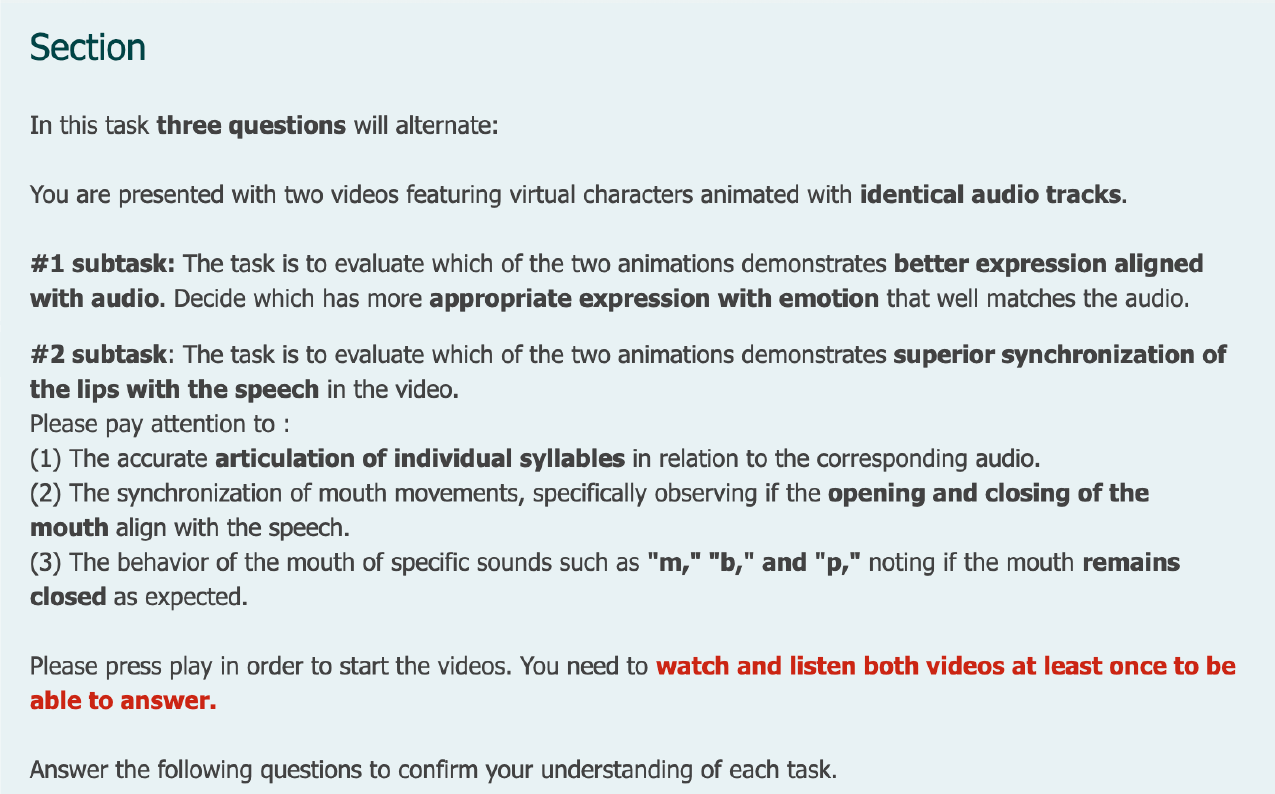}
   \vspace{-5pt}
    \caption{Instruction for User Study}
   \vspace{-5pt}
    \label{fig:instruction for user study}
\end{figure*}

\begin{figure*}[h]
    \centering
    \includegraphics[width=0.9\linewidth]{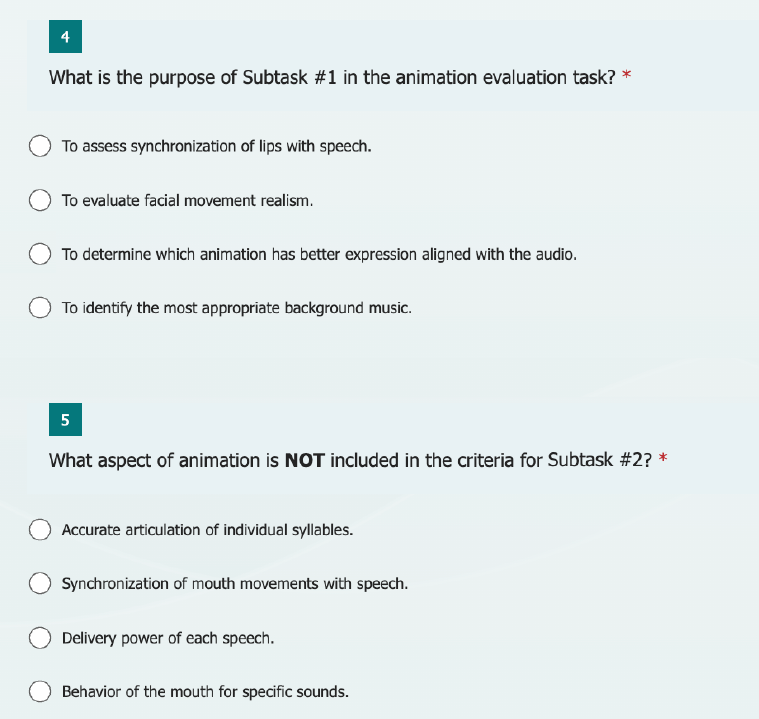}
   \vspace{-5pt}
    \caption{Confirm question for User Study}
   \vspace{-5pt}
    \label{fig:confirm for user study}
\end{figure*}

\begin{figure*}[h]
    \centering
    \includegraphics[width=0.9\linewidth]{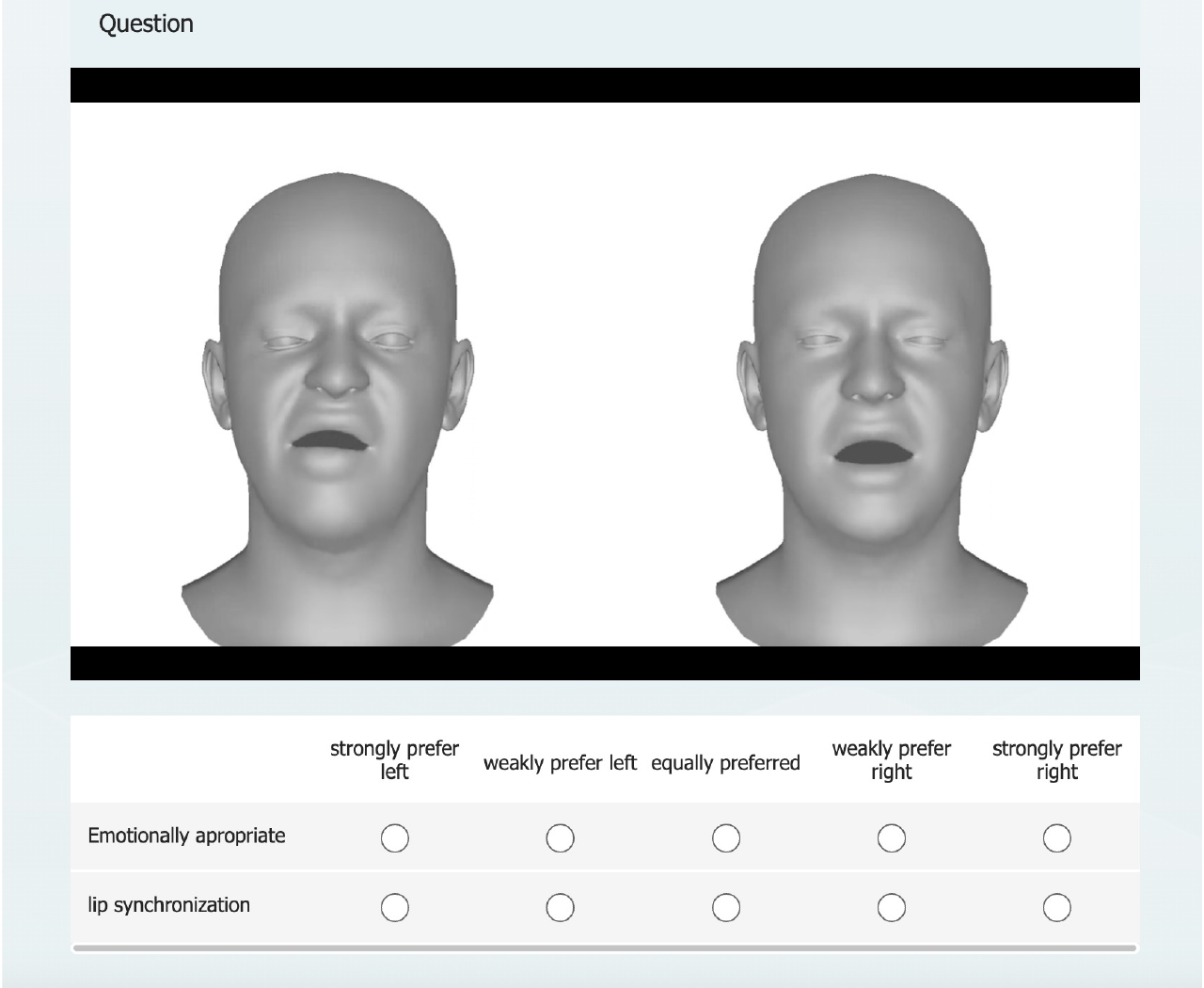}
   \vspace{-5pt}
    \caption{UI for User Study. Each part has a number of questions for one model for comparison, or an ablation of the DEEPTalk model.}
   \vspace{-5pt}
    \label{fig:UI for user study}
\end{figure*}





\end{document}